\pdfoutput=1

\documentclass[11pt]{article}

\usepackage[final]{acl}

\usepackage{times}
\usepackage{latexsym}

\usepackage[T1]{fontenc}

\usepackage[utf8]{inputenc}

\usepackage{microtype}

\usepackage{inconsolata}

\usepackage{longtable}
\usepackage{xcolor}
\usepackage{colortbl}
\usepackage{multicol}
\usepackage{booktabs}
\usepackage{makecell}
\usepackage{amsmath, amssymb}
\usepackage{multirow}
\usepackage{mathtools}
\usepackage[inline]{enumitem}
\usepackage{subcaption}
\usepackage{float}
\usepackage{graphicx}
\usepackage{caption} 
\usepackage{upgreek}
\usepackage{seqsplit}
\usepackage{color, soul}
\usepackage{arydshln}
\usepackage{xspace}
\usepackage{comment}
\usepackage{url}
\usepackage{relsize}

\usepackage{amsmath,amsfonts,bm}

\def\eqref#1{equation~\ref{#1}}

\def\1{\bm{1}}

\def\vf{{\bm{f}}}

\def\vk{{\bm{k}}}

\def\vq{{\bm{q}}}

\def\vt{{\bm{t}}}

\def\vv{{\bm{v}}}

\def\vy{{\bm{y}}}

\def\mF{{\bm{F}}}

\def\mV{{\bm{V}}}

\DeclareMathAlphabet{\mathsfit}{\encodingdefault}{\sfdefault}{m}{sl}
\SetMathAlphabet{\mathsfit}{bold}{\encodingdefault}{\sfdefault}{bx}{n}

\newcommand{\comb}[2]{{}_{#1}\mathrm{C}_{#2}}

\usepackage{caption}
\usepackage{subcaption}
\newcommand{\ours}{VideoRAG\xspace}
\title{VideoRAG: Retrieval-Augmented Generation over Video Corpus}

\author{
    Soyeong Jeong$^{1*}$ \; 
    Kangsan Kim$^{1*}$ \;
    Jinheon Baek$^{1*}$ \;
    Sung Ju Hwang$^{1,2}$ \\
    KAIST$^{1}$ \;\; DeepAuto.ai$^{2}$ \\
    \texttt{\{starsuzi, kksan07, jinheon.baek, sungju.hwang\}@kaist.ac.kr}
}

\begin{document}
\maketitle
\def\thefootnote{*}\footnotetext{Equal contribution}\def\thefootnote{\arabic{footnote}}

\begin{abstract}

Retrieval-Augmented Generation (RAG) is a powerful strategy for improving the factual accuracy of models by retrieving external knowledge relevant to queries and incorporating it into the generation process. However, existing approaches primarily focus on text, with some recent advancements considering images, and they largely overlook videos, a rich source of multimodal knowledge capable of representing contextual details more effectively than any other modality. Also, while very recent studies explore the use of videos in response generation, they either predefine query-associated videos without retrieval or convert videos into textual descriptions, losing multimodal richness. To tackle these, we introduce VideoRAG, a novel framework that not only dynamically retrieves videos based on their relevance with queries but also utilizes both visual and textual information. The operation of VideoRAG is powered by recent Large Video Language Models (LVLMs), which enable the direct processing of video content to represent it for retrieval and the seamless integration of retrieved videos jointly with queries for response generation. Also, inspired by that the context size of LVLMs may not be sufficient to process all frames in extremely long videos and not all frames are equally important, we introduce a video frame selection mechanism to extract the most informative subset of frames, along with a strategy to extract textual information from videos (as it can aid the understanding of video content) when their subtitles are not available. We experimentally validate the effectiveness of VideoRAG, showcasing that it is superior to relevant baselines. Our code is available at \url{https://github.com/starsuzi/VideoRAG}.

\end{abstract}
\section{Introduction}

\begin{figure}
    \centering
    \includegraphics[width=0.975\linewidth]{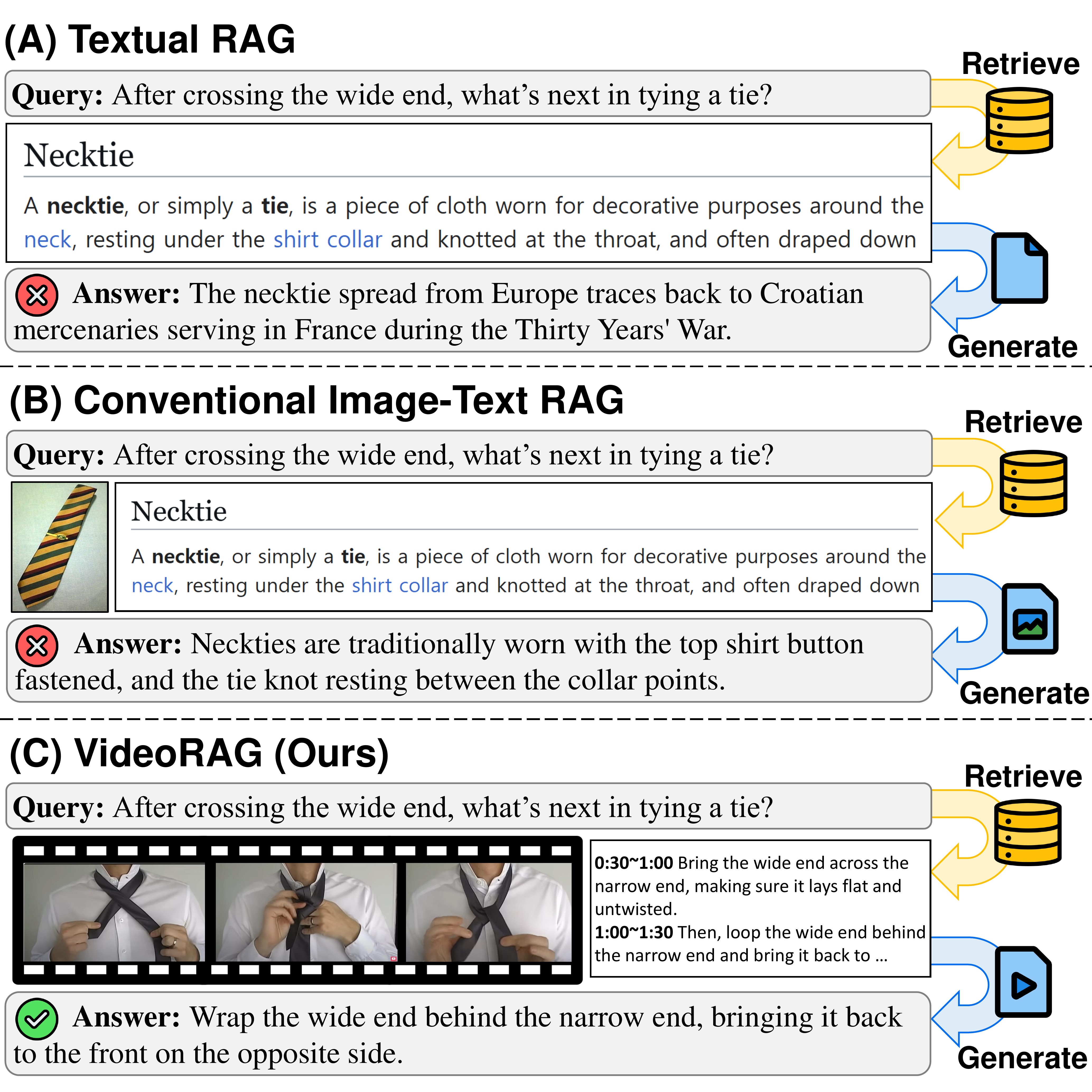}
    \vspace{-0.1in}
    \caption{\small Illustration of existing and the proposed RAG scenarios. (A) Textual RAG retrieves documents (relevant to queries) from a text corpus and incorporates them when generating answers. (B) Conventional image-text multimodal RAG extends retrieval to include static images. (C) \textsc{VideoRAG} (ours) further extends the external knowledge source to videos.}
    \label{fig:fig_concept}
    \vspace{-0.175in}
\end{figure}

Recently, large foundation models, such as large language models and their extension to the vision modality called large vision-language models, have become the standard for addressing diverse tasks due to their remarkable capabilities~\cite{GPT-4, llavaonevision, qwen2, NVLM}. In particular, these models, trained on extensive textual and multimodal corpora, encode vast amounts of knowledge within their large-scale parameters. However, they are still prone to generating factually incorrect outputs, as their parametric knowledge can be inaccurate or outdated~\cite{rag, ralm}. This limitation highlights the need for incorporating knowledge from external knowledge sources, with Retrieval-Augmented Generation (RAG) emerging as an essential mitigator for it. Specifically, RAG typically operates by retrieving query-relevant information and then generating answers grounded in the retrieved content~\cite{RAGTruth, DBLP:conf/naacl/AyalaB24}. 

However, while existing RAG approaches have been widely adopted for various real-world applications, they have primarily focused on retrieving and incorporating textual content \cite{ralm, adaptive-rag}, with only recent attempts beginning to explore images (or text-image pairs) as the additional source of external knowledge~\cite{visRAG, DBLP:journals/corr/abs-2410-21943}. On the other hand, we argue that there remains a rapidly expanding yet underutilized medium, called videos, which provides unparalleled multimodal richness and might be a compelling resource for augmenting the knowledge landscape of current RAG systems. Specifically, videos combine temporal dynamics, spatial details, and multimodal cues, which collectively enable them to capture complex processes, context-dependent interactions, and non-verbal signals that static modalities (e.g., text and images) often fail to convey. Moreover, given the increasing popularity of video-sharing platforms (such as YouTube), the availability of diverse, high-quality video data has grown, ranging from educational tutorials and scientific demonstrations to personal experiences and real-time events, all of which may be useful when formulating responses to user queries. 

A few recent studies have started considering video content to handle user queries; however, they have limitations. For instance, some assume that videos relevant to queries are already known and instead focus on identifying query-relevant frames within that specified video~\cite{videorag,drvideo}. While effective in scenarios where the relevant video is explicitly provided, it is suboptimal for more general-use cases, where users expect systems to dynamically identify and retrieve videos to provide answers. On the other hand, other studies handle videos by converting them into textual formats, such as subtitles, and utilizing these textual representations under off-the-shelf text-based RAG pipelines~\cite{iRAG,OmAgent}. However, while this text-only strategy may offer a convenient workaround, it inherently sacrifices the multimodal richness of video data by discarding critical information, such as temporal dynamics captured in the visual context, during the conversion process. For example, consider a query: ``How does the expression of the dog change when it is angry?''. While textual transcriptions might describe the dog's barking or growling, they fail to capture visual cues (baring teeth, raised hackles, or narrowed eyes), which are needed for accurately interpreting the emotional state of the dog and subsequently formulating the answer to the query.

\begin{figure*}
    \centering
    \includegraphics[width=0.975\linewidth]{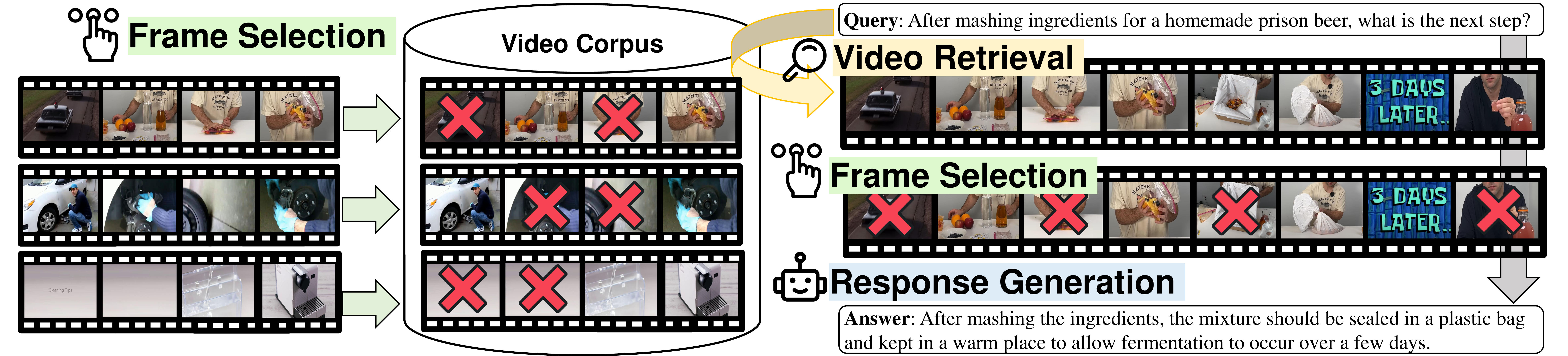}
    \vspace{-0.075in}
    \caption{\small Illustration of the overall pipeline of our VideoRAG, which selects informative frames for retrieval and generation.}
    \label{fig:fig_concept_method}
    \vspace{-0.1in}
\end{figure*}

To address the aforementioned limitations, we introduce a novel framework, called VideoRAG, which aims to offer another fruitful angle to existing RAG frameworks by enabling a more comprehensive utilization of video content for its holistic retrieval and incorporation (See Figure~\ref{fig:fig_concept}). Specifically, in response to queries, the proposed VideoRAG retrieves relevant videos from a large video corpus but also integrates both visual and textual elements into the answer-generation process. Also, we operationalize this by harnessing the advanced capabilities of recent Large Video Language Models (LVLMs), which are capable of directly processing video content, consisting of visual and textual information, within the unified framework, thereby more effectively capturing its multimodal richness. 

However, there exist a couple of remaining challenges in integrating videos into RAG frameworks. First, videos are inherently long and redundant, oftentimes making it infeasible for LVLMs to process all frames due to their limited context capacity as well as unnecessary since not all frames contribute meaningfully for retrieval and generation. To address this, we introduce a frame selection model that is trained to extract the most informative subset of frames to maximize retrieval and generation performance. Also, we observe that, while the joint utilization of visual and textual features is needed for the effective representation of videos and subsequently their retrieval, the textual descriptions of videos (e.g., subtitles) are oftentimes not available. To tackle this, we further present a simple yet effective mitigation strategy that utilizes automatic speech recognition techniques to generate textual transcripts from videos, allowing us to leverage both visual and textual modalities for every video. 

To validate the effectiveness of VideoRAG, we conduct experiments by using overlapping queries from the WikiHowQA dataset~\cite{WikiHowQA} (consisting of query-answer pairs) and the HowTo100M dataset~\cite{HowTo100M} (including query-video pairs without answers). Also, based on this, we automatically collect the dataset for RAG over videos and then evaluate models on it. Then, the experimental results show the significant performance improvement of the proposed VideoRAG framework over relevant baselines, demonstrating the efficacy of leveraging videos for RAG.

\section{Method}

We present VideoRAG that retrieves query-relevant videos and generates responses grounded in them. 

\subsection{Preliminaries}
\label{sec:preliminaries}
We begin with describing RAG and LVLMs.

\paragraph{Retrieval-Augmented Generation}
RAG aims to enhance the capabilities of foundation models by grounding their outputs in external knowledge retrieved from the external knowledge source, such as Wikipedia, which consists of two main components: retrieval and generation modules. Formally, given a query $\vq$, RAG retrieves a set of documents (or knowledge elements) $\mathcal{K} = \{ \vk_1, \vk_2, \dots, \vk_k \}$ from an external corpus $\mathcal{C}$ ($\mathcal{K} \subseteq \mathcal{C}$) based on their relevance with $\vq$ using a retrieval module, which can be formalized as follows: $\mathcal{K} = \texttt{Retriever}(\vq, \mathcal{C})$. Here, the query $\vq$ and knowledge $\vk$ are represented as a sequence of tokens $\vq = [q_1, q_2, \dots, q_i]$ and $\vk = [k_1, k_2, \dots, k_j]$. Also, during retrieval, the relevance between the query and each knowledge element within the corpus is determined by the scoring function, defined as follows: $\texttt{Sim}(\vq, \vk)$, which typically measures their representational similarity over the embedding space. In the subsequent generation step, the retrieved knowledge elements are then used as additional input to the generation module, to augment the query to produce an answer $\vy$, as follows: $\vy = \texttt{Model}(\vq, \mathcal{K})$, where $\texttt{Model}$ is typically implemented as the foundation model, such as LLMs. We note that, unlike existing RAG that focuses mainly on retrieving and incorporating textual content (or, in some recent cases, extra static images), we explore the extension toward videos.

\paragraph{Large Video Language Models}
On top of the extensive language understanding capabilities of LLMs, LVLMs are designed to handle and incorporate the features from video content, including temporal, spatial, and multimodal information, within the unified token processing framework. Formally, let us denote a video $\mV$ as a sequence of visual frames: $\mV = [\vv_1, \vv_2, \dots, \vv_n]$ and its associated textual data (such as subtitles, or any other textual information such as the video-specific query) $\vt$ as a sequence of tokens: $\vt = [t_1, t_2, \dots, t_m]$. Then, the typical LVLM, denoted as $\texttt{LVLM}$, enables the joint processing of these multimodal inputs by employing two specialized components: a vision encoder and a text encoder. Specifically, the vision encoder processes the sequence of video frames $\mV$ (which can span multiple videos), resulting in a sequence of visual feature embeddings (or visual tokens): $\mF_{\texttt{visual}} = \texttt{VisionEncoder}(\mV)$. Concurrently, the text encoder processes the given textual information $\vt$ to generate corresponding feature embeddings: $\mF_{\texttt{text}} = \texttt{TextEncoder}(\vt)$. Then, the overall process to obtain the video representation (with the goal of capturing both visual and textual features) can be denoted as follows: $\vf_{\texttt{video}} = \texttt{LVLM}(\mV, \vt)$. Traditionally, $\vf_{\texttt{video}}$ is obtained by the simple interpolation of the visual and textual representations: $\vf_{\texttt{video}} = \alpha \cdot \mF_{\texttt{text}} + (1-\alpha) \cdot \mF_{\texttt{visual}}$~\cite{VideoCLIP}, and, more recently, it can be done by further jointly processing the visual and textual embeddings through several LVLM layers (that sit on top of existing LLMs)~\cite{llavanext-video}, which allows the model to learn a more effective representation and continue generating the next sequence of tokens (for example, an answer to a query).

\subsection{VideoRAG}
We now turn to introduce our VideoRAG, which extends the existing RAG paradigm by leveraging the video corpus as the external knowledge source. 

\paragraph{Video Retrieval}
The initial step to operationalize RAG over the video corpus is to implement video retrieval, whose goal is to identify query-relevant videos $\mathcal{V} = \{ \mV_1, \mV_2, \dots, \mV_k \}$ from the corpus $\mathcal{C}$, consisting of a large number of videos, as follows: $\mathcal{V} = \texttt{Retriever}(\vq, \mathcal{C})$. Recall that this retrieval process involves calculating the similarity between the query $\vq$ and each knowledge element (which is video $\mV$ in our case) to determine their relevance. To achieve this, we first forward the video $\mV$ (composed of image frames and, if available, subtitles) as well as the query $\vq$ (without visual information) into $\texttt{LVLM}$, to obtain their representations $\vf_{\texttt{query}}$ and $\vf_{\texttt{video}}$. After that, the relevance is computed based on their representation-level similarity, for example, using a cosine similarity, and the top-$k$ videos with the highest similarity scores are retrieved. 

\paragraph{Video-Augmented Response Generation}
After the retrieval of query-relevant videos is done, the next step is to incorporate the retrieved videos into the answer generation process, to formulate the answer grounded in them. To operationalize this, we first concatenate frames of each retrieved video with its associated textual data (e.g., subtitles), then concatenate these multimodal pairs across all videos retrieved, and lastly append the user query, as follows: $[\mV_1, \vt_1, \dots, \mV_k, \vt_k, \vq]$. Then, this input is forwarded into $\texttt{LVLM}$, which enables the joint processing of the combined visual, textual, and query-specific information, to generate the response while capturing their multimodal richness and dynamics.

\subsection{Frame Selection for VideoRAG}
\label{sec:frame_selection}
Unlike conventional RAG with text or images, incorporating videos into RAG presents an additional challenge: some videos contain a large number of visual frames, making it inefficient to process them all (and sometimes impractical due to the limited context size of LVLMs). As a simple workaround, a common approach is to uniformly sample frames; however, this method risks discarding key information while retaining redundant or irrelevant frames, leading to suboptimal retrieval and response generation when augmented with suboptimal frames. 

\paragraph{Adaptive Frame Selection}
To overcome these limitations, we introduce an adaptive frame selection strategy, whose objective is to extract the most informative and computationally feasible subset of frames. Let $\texttt{Comb}(\cdot)$ represent a selection function that randomly samples a subset of $m$ frames from total $n$ frames within the video based on the combination, and let $f(\cdot)$ be a scoring function that evaluates and assigns a relevance score to these selected frames. Then, during retrieval, the frame selection operation for the given video $\mV$ is denoted as follows: $\tilde{\mV} = \arg \max_{\mV' \in \texttt{Comb}(\mV, m)} f(\mV')$, which is extended to $\tilde{\mV} = \arg \max_{\mV' \in \texttt{Comb}(\mV, m)} f(\mV', \vq)$ for generation, where $\tilde{\mV}$ is the optimal subset. The distinction between retrieval and generation arises because retrieval operates over a large video corpus $\mathcal{C}$, making exhaustive query-based processing infeasible, whereas in generation, the top-$k$ retrieved videos allow for query-guided frame selection (i.e., enabling the use of different frames for different queries even if the retrieved video is the same).

\paragraph{Frame Space Reduction with Clustering}
While the adaptive frame selection strategy enables the use of the most effective subset of frames for RAG, the combinatorial space of possible frame subsets (obtained from $\texttt{Comb}$) remains prohibitively large. For instance, selecting 32 frames from a video of 1000 frames results in more than $10^{60}$ possible combinations, making exhaustive search impossible. To address this, we reduce the frame selection space by extracting representative samples via $k$-means++ clustering. Specifically, we cluster all frames into $k$ groups and, from each of the $k$ clusters, we select the frame closest to its centroid. After that, we constrain the frame selection process to operate within this reduced set; for example, with $k=64$, the search space is drastically reduced to $\comb{64}{32}$ from $\comb{1000}{32}$, making it computationally feasible while preserving the diversity of selected frames\footnote{In inference, evaluating all possible combinations from this reduced set might still be computationally expensive; thus, we further perform random sampling over them.}. 

\paragraph{Operationalizing Frame Selection}
Notably, the design of $f$ to score the selected frame is flexible, allowing us to use any models capable of processing visual features (and textual features particularly for generation), such as CLIP~\cite{clip}. Also, we collect examples for training $f$, by performing retrieval and generation with randomly selected frames (from possible combinations), and then labeling them as true or false based on their success, from which we use the conventional loss functions (such as cross-entropy) for optimization. We provide more details on it in Appendix~\ref{sec:appen_frame_selection}.

\subsection{Auxiliary Text Generation}
In both the retrieval and generation steps, the inclusion of video-associated textual data, such as subtitles, can play a crucial role in enhancing video representation since it provides additional context and semantic cues that complement the visual content. However, not every video in the corpus comes with subtitles since they require additional annotations. Therefore, for such videos, we propose generating auxiliary textual data by extracting audio from the video and converting it into text using off-the-shelf automatic speech recognition techniques. Formally, given a video $\mV$, this process can be formalized as follows: $\vt_{\texttt{aux}} = \texttt{AudioToText}(\texttt{Audio}(\vv))$, where $\texttt{Audio}(\mV)$ extracts the audio track from the video, and $\texttt{AudioToText}$ converts the extracted audio signal into textual content. Therefore, for those videos without subtitles, auxiliary text $\vt_{\texttt{aux}}$ can be used in place of $\vt$ in both the retrieval and generation steps.

\section{Experiment}
We now describe experimental setup and results.

\begin{table*}[t!]
\newcommand{\oracle}[1]{\textcolor{gray}{#1}}   
\caption{\small Overall RAG results across four metrics. The best results are highlighted in \textbf{bold}, and the second-best results are highlighted with \underline{underline}. Note that the \textsc{Oracle} setting (that uses ideal retrieval results) is not comparable to others.}
\vspace{-0.1in}
\label{tab:main}
\small
\centering
\renewcommand{\arraystretch}{1.0}
\renewcommand{\tabcolsep}{2mm}
\resizebox{\linewidth}{!}{%
    \begin{tabular}{p{5mm}l cccc cccc}
    \toprule
    
    & & \multicolumn{4}{c}{\bf WikiHowQA with HowTo100M} & \multicolumn{4}{c}{\bf Synthetic QA with HowTo100M} \\
    \cmidrule(l{2pt}r{2pt}){3-6} \cmidrule(l{2pt}r{2pt}){7-10} 
    
    & \textbf{Methods} & \textbf{ROUGE-L} & \textbf{BLEU-4} & \textbf{BERTScore} & \textbf{G-Eval} & \textbf{ROUGE-L} & \textbf{BLEU-4} & \textbf{BERTScore} & \textbf{G-Eval} \\
    

    \midrule
    \midrule
    \multirow{9}{*}[-0.25em]{\rotatebox[origin=c]{90}{\textbf{LLaVA-Video (7B)}}}
    & \textbf{\textsc{Naïve}} & 14.08 & 1.352 & 83.43 & 1.579 & 10.68 & 1.574 & 84.51 & 1.634 \\
    
    & \textbf{\textsc{TextRAG (BM25)}} & 17.22 & 2.327 & 84.66 & 1.633 & 14.70 & 2.382 & 86.03 & 1.681 \\

    & \textbf{\textsc{TextRAG (DPR)}} & 16.65 & 2.173 & 84.61 & 1.591  & 14.58 & 2.397 & 85.85 & 1.686 \\
    
    & \textbf{\textsc{TextImageRAG}} & 22.43 & 4.222 & 86.88 & 2.022 & 25.19 & 6.149 & 88.56 & 2.175 \\

    & \textbf{\textsc{TextVideoRAG}} & 22.81 & 4.388 & 86.97 & 1.979 & 23.41 & 5.435 & 88.40 & 2.278 \\

    \noalign{\vskip 0.25ex}\cdashline{2-10}\noalign{\vskip 0.75ex}

    & \cellcolor{blue!5}\textbf{\textsc{VideoRAG-V}} & \cellcolor{blue!5}\textbf{24.95} & \cellcolor{blue!5}\underline{5.080} & \cellcolor{blue!5}\underline{87.85} & \cellcolor{blue!5}\underline{2.140} & \cellcolor{blue!5}\underline{29.38} & \cellcolor{blue!5}\underline{7.530} & \cellcolor{blue!5}\textbf{89.77} & \cellcolor{blue!5}\textbf{2.479} \\

    & \cellcolor{blue!5}\textbf{\textsc{VideoRAG-VT}} & \cellcolor{blue!5}\underline{24.93} & \cellcolor{blue!5}\textbf{5.276} & \cellcolor{blue!5}\textbf{87.92} & \cellcolor{blue!5}\textbf{2.142} & \cellcolor{blue!5}\textbf{29.74} & \cellcolor{blue!5}\textbf{8.043} & \cellcolor{blue!5}\underline{89.72} & \cellcolor{blue!5}\underline{2.476} \\
    
    \noalign{\vskip 0.25ex}\cdashline{2-10}\noalign{\vskip 0.75ex}
    
    & \textbf{\textsc{Oracle-V}} & 26.19 & 5.480 & 88.41 & 2.225 &  32.16 & 8.769 & 90.34 & 2.884 \\
    
    & \textbf{\textsc{Oracle-VT}} & 25.37 & 5.237 & 87.95 & 2.166 & 32.31 & 8.885 & 90.46 & 2.938 \\

    \midrule
    \midrule
    \multirow{12}{*}[1.25em]{\rotatebox[origin=c]{90}{\textbf{InternVL2.5 (8B)}}} 
    & \textbf{\textsc{Naïve}} & 16.54 & 1.859 & 84.30 & 1.720 & 12.60 & 2.381 & 85.12 & 1.725 \\
    
    & \textbf{\textsc{TextRAG (BM25)}} & 17.41 & 2.275 & 84.89 & 1.552 & 26.66 & 6.760 & 88.48 & 1.938 \\

    & \textbf{\textsc{TextRAG (DPR)}} & 17.21 & 2.077 & 84.84 & 1.563 & 26.72 & 6.579 & 88.56 & 1.917 \\
    
    & \textbf{\textsc{TextImageRAG}} & 22.39 & 3.917 & 86.91 & \underline{1.904} & 27.65 & 7.187 & 88.99 & 2.176 \\

    & \textbf{\textsc{TextVideoRAG}} & 19.88 & 3.199 & 85.81 & 1.686 & 26.36 & 6.542 & 88.68 & 1.983 \\

    \noalign{\vskip 0.25ex}\cdashline{2-10}\noalign{\vskip 0.75ex}

    & \cellcolor{blue!5}\textbf{\textsc{VideoRAG-V}} & \cellcolor{blue!5}\textbf{25.11} & \cellcolor{blue!5}\underline{4.243} & \cellcolor{blue!5}\textbf{88.15} & \cellcolor{blue!5}1.863 & \cellcolor{blue!5}\textbf{33.68} & \cellcolor{blue!5}\underline{9.454} & \cellcolor{blue!5}\textbf{90.29} & \cellcolor{blue!5}\textbf{2.452} \\

    & \cellcolor{blue!5}\textbf{\textsc{VideoRAG-VT}} & \cellcolor{blue!5}\underline{23.75} & \cellcolor{blue!5}\textbf{4.271} & \cellcolor{blue!5}\underline{87.42} & \cellcolor{blue!5}\textbf{1.906} & \cellcolor{blue!5}\underline{32.90} & \cellcolor{blue!5}\textbf{9.572} & \cellcolor{blue!5}\underline{90.14} & \cellcolor{blue!5}\underline{2.427} \\
    
    \noalign{\vskip 0.25ex}\cdashline{2-10}\noalign{\vskip 0.75ex}
    
    & \textbf{\textsc{Oracle-V}} & 25.59 & 4.318 & 88.29 & 1.958 & 35.21 & 10.57 & 90.70 & 2.813 \\
    
    & \textbf{\textsc{Oracle-VT}} & 24.60 & 4.421 & 87.70 & 2.002 & 34.99 & 10.69 & 90.68 & 2.820 \\

    \midrule
    \midrule
    
    \multirow{12}{*}[1.25em]{\rotatebox[origin=c]{90}{\textbf{Qwen2.5-VL (3B)}}} 
    & \textbf{\textsc{Naïve}} & 17.96 & 2.077 & 84.97 & 1.765 & 15.05 & 2.729 & 86.13 & 1.843 \\
    
    & \textbf{\textsc{TextRAG (BM25)}} & 19.65 & 2.989 & 85.41 & 1.721 & 19.70 & 3.911 & 86.88 & 1.877 \\
    
    & \textbf{\textsc{TextRAG (DPR)}} & 19.45 & 2.863 & 85.38 & 1.708 & 19.04 & 3.903 & 86.77 & 1.831 \\
    
    & \textbf{\textsc{TextImageRAG}} & 20.66 & 3.327 & 85.80 & 1.838 & 20.36 & 4.298 & 87.11 & 1.931 \\
    
    & \textbf{\textsc{TextVideoRAG}} & 22.18 & \underline{4.180} & 86.56 & 1.821 & 24.29 & 5.722 & 88.37 & 2.156 \\
    
    \noalign{\vskip 0.25ex}\cdashline{2-10}\noalign{\vskip 0.75ex}

    & \cellcolor{blue!5}\textbf{\textsc{VideoRAG-V}} & \cellcolor{blue!5}\textbf{23.24} & \cellcolor{blue!5}3.963 & \cellcolor{blue!5}\textbf{87.13} & \cellcolor{blue!5}\textbf{1.899} & \cellcolor{blue!5}\underline{26.28} & \cellcolor{blue!5}\underline{5.998} & \cellcolor{blue!5}\underline{88.97} & \cellcolor{blue!5}\underline{2.258} \\

    & \cellcolor{blue!5}\textbf{\textsc{VideoRAG-VT}} & \cellcolor{blue!5}\underline{23.22} & \cellcolor{blue!5}\textbf{4.531} & \cellcolor{blue!5}\underline{87.00} & \cellcolor{blue!5}\underline{1.876} & \cellcolor{blue!5}\textbf{27.54} & \cellcolor{blue!5}\textbf{7.279} & \cellcolor{blue!5}\textbf{89.11} & \cellcolor{blue!5}\textbf{2.274} \\
    
    \noalign{\vskip 0.25ex}\cdashline{2-10}\noalign{\vskip 0.75ex}
    
    & \textbf{\textsc{Oracle-V}} & 21.53 & 3.156 & 86.05 & 1.912 & 26.82 & 6.683 & 88.96 & 2.515 \\
    
    & \textbf{\textsc{Oracle-VT}} & 24.37 & 4.811 & 87.43 & 1.994 & 29.76 & 7.721 & 89.56 & 2.566 \\

    \bottomrule
    
    \end{tabular}
}
\vspace{-0.025in}
\end{table*}

\subsection{Experimental Setup}

\paragraph{Datasets}
We evaluate VideoRAG in question answering tasks, following the convention for validating RAG approaches~\cite{self-rag, adaptive-rag}. First of all, we use WikiHowQA~\cite{WikiHowQA}, which offers a wide range of instructional questions extracted from the WikiHow webpage\footnote{\url{https://www.wikihow.com/Main-Page}}, with human-written, high-quality ground truths. Also, for the video corpus, we utilize HowTo100M~\cite{HowTo100M}, a comprehensive collection of instruction videos sourced from YouTube, further associated with queries from WikiHow based on their search results. In addition, for a comprehensive evaluation, we automatically generate query-answer pairs over HowTo100M (See Appendix~\ref{sec:appen_syn_data_gen}) and evaluate performance on them.

\paragraph{Baselines and Our Model}
We compare \ours against four different baselines, as follows:
\begin{enumerate*}[itemsep=0.0mm, parsep=1pt, leftmargin=*]
    \item \textbf{\textsc{Naïve}} -- which generates answers from queries without additional context;
    \item \textbf{\textsc{TextRAG (BM25)}} -- which is a text-based RAG model, retrieving documents (from Wikipedia) based on their relevance with queries through BM25~\cite{bm25} and generating answers grounded in them; 
    \item \textbf{\textsc{TextRAG (DPR)}} -- which is a text-based RAG similar to \textsc{TextRAG (BM25)} but performs retrieval with DPR~\cite{dpr};
    \item \textbf{\textsc{TextImageRAG}} -- which follows conventional text-image multimodal RAG approaches~\cite{MuRAG, DBLP:conf/icml/YasunagaAS0LLLZ23}, retrieving a pair of query-relevant textual document and image, and utilizing them for generation;
    \item \textbf{\textsc{TextVideoRAG}} -- which follows the previous video-based RAG methods~\cite{iRAG,OmAgent}, which first represent videos as their textual descriptions (e.g., captions or transcripts) and utilize only those textual information in retrieval and generation;
    \item \textbf{\textsc{VideoRAG}} -- which is our model having two variants: \textbf{\textsc{VideoRAG-V}} that exclusively utilizes video frames as context to provide visual grounding for generation, and \textbf{\textsc{VideoRAG-VT}} that jointly utilizes video frames and textual transcripts. 
\end{enumerate*}
In addition, to estimate the room for performance gains, we include an oracle version of \textsc{VideoRAG}, which directly uses the ground-truth video pre-associated with the query labeled in HowTo100M, instead of using retrieval outcomes.

\paragraph{Evaluation Metrics}
We use the following metrics: 
\textbf{1) ROUGE-L} measures the longest common subsequence between the generated answer and the ground truth~\cite{ROUGE};
\textbf{2) BLEU-4} calculates the overlap of n-grams (up to 4) between the generated and reference answers~\cite{BLEU};
\textbf{3) BERTScore} measures the semantic alignment between the generated and reference answers~\cite{BERTScore} by extracting their embeddings from BERT~\cite{BERT} and calculating their similarity;
\textbf{4) G-Eval} leverages the evaluation capabilities of LLMs~\cite{G-Eval}, where we prompt the GPT-4o-mini to rate the generated answer in comparison to the reference on a 5-point Likert scale, with a prompt provided in Table~\ref{tab:prompt:geval}.

\begin{figure*}[t!]
    \centering
    \vspace{-0.1in}
    \begin{minipage}{0.33\textwidth}
        \centering
        \renewcommand{\arraystretch}{1.25}
        \setlength{\tabcolsep}{6pt}
        \vspace{0.1in}
        \resizebox{\linewidth}{!}{%
            \begin{tabular}{lccc}
            \toprule
            \textbf{Features} & \textbf{R@1} & \textbf{R@5} & \textbf{R@10} \\
            \midrule
            \midrule
            \textbf{Visual} & 0.054 & 0.193 & 0.288 \\
            \textbf{Textual} & 0.088 & 0.302 & 0.388 \\
            \textbf{Ensemble} & \textbf{0.103} & \textbf{0.311} & \textbf{0.442} \\ \bottomrule
            \end{tabular}
        }
        \vspace{-0.06in}
        \captionof{table}{\small Retrieval results, where we use visual features alone, textual features alone, or an ensemble of their features.}
        \label{tab:feat_ensemble}
    \end{minipage}
    \hfill
    \begin{minipage}{0.30\linewidth}
        \centering
        \includegraphics[width=0.95\columnwidth]{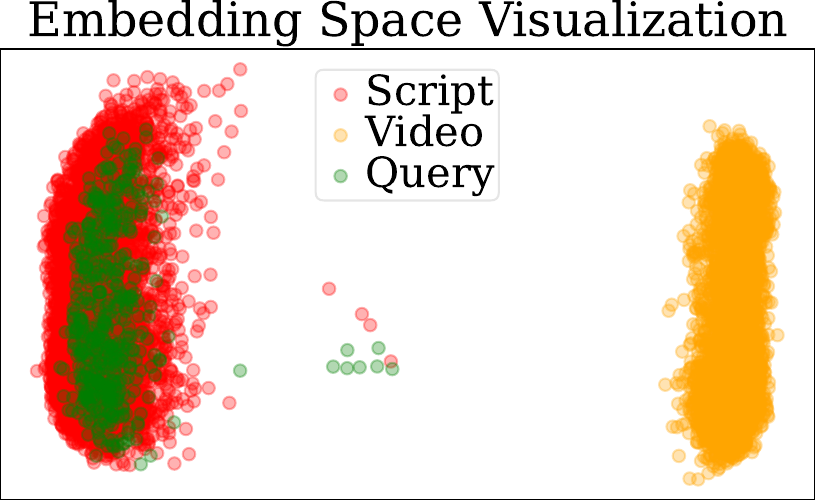}
        \vspace{-0.06in}
        \caption{\small Visualization of latent space of features across modalities with Principal Component Analysis (PCA).}
        \label{fig:pca}
    \end{minipage}
    \hfill
    \begin{minipage}{0.33\linewidth}
        \centering
        \includegraphics[width=0.95\columnwidth]{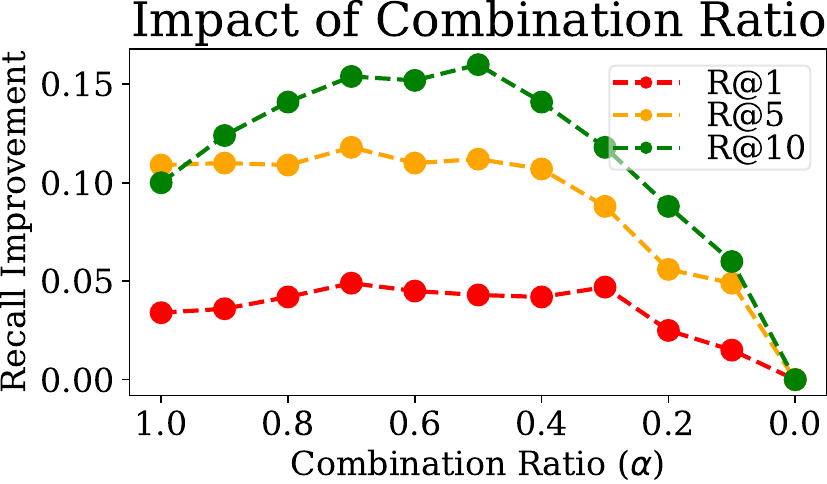}
        \vspace{-0.1in}
        \caption{\small Impact of varying the interpolation ratio between textual and visual features on the video retrieval performance.}
        \label{fig:alpha}
    
    \end{minipage}
    \label{fig:video_ret}
    \vspace{-0.125in}
\end{figure*}

\paragraph{Implementation Details}
We consider multiple LVLMs: LLaVA-Video of 7B, InternVL 2.5 of 8B, and Qwen-2.5-VL of 3B parameters for generation~\cite{llavanext-video, internvl2.5, Qwen2.5-VL}, alongside InternVideo2~\cite{internvideo2} for retrieval (please see Appendix~\ref{sec:appen_model_choice} for details on model choice). For efficiency, we use 4 frames per video for retrieval, while we use 32 frames (or all frames if the video is shorter than 32 seconds, sampled at 1 fps) for generation. In auxiliary text generation, we use Whisper~\cite{whisper}.

\subsection{Experimental Results and Analyses}
We now present results and various analyses.

\paragraph{Main Results}
We provide main results in Table~\ref{tab:main}, showcasing the performance of different models with varying types of retrieved knowledge. First, we find that all RAG models clearly outperform the \textsc{Naïve} baseline, reaffirming the critical role of external knowledge in enhancing the factual accuracy of generated responses. Also, among these, our \textsc{\ours} achieves the best performance, significantly surpassing conventional textual, text-image, or text-video RAG  baselines. This improvement corroborates our hypothesis that video content is a useful resource for RAG since it provides richer and more detailed information than other modalities. Lastly, the smaller performance gap between \textsc{\ours-V} and \textsc{\ours-VT} suggests that much of the necessary information required for answer generation is effectively encapsulated within visual features of videos, which inherently include information conveyed through textual descriptions.

\begin{figure*}[t!]
    \centering
    \hfill
    \begin{minipage}{0.35\linewidth}
        \centering
        \includegraphics[width=0.975\linewidth]{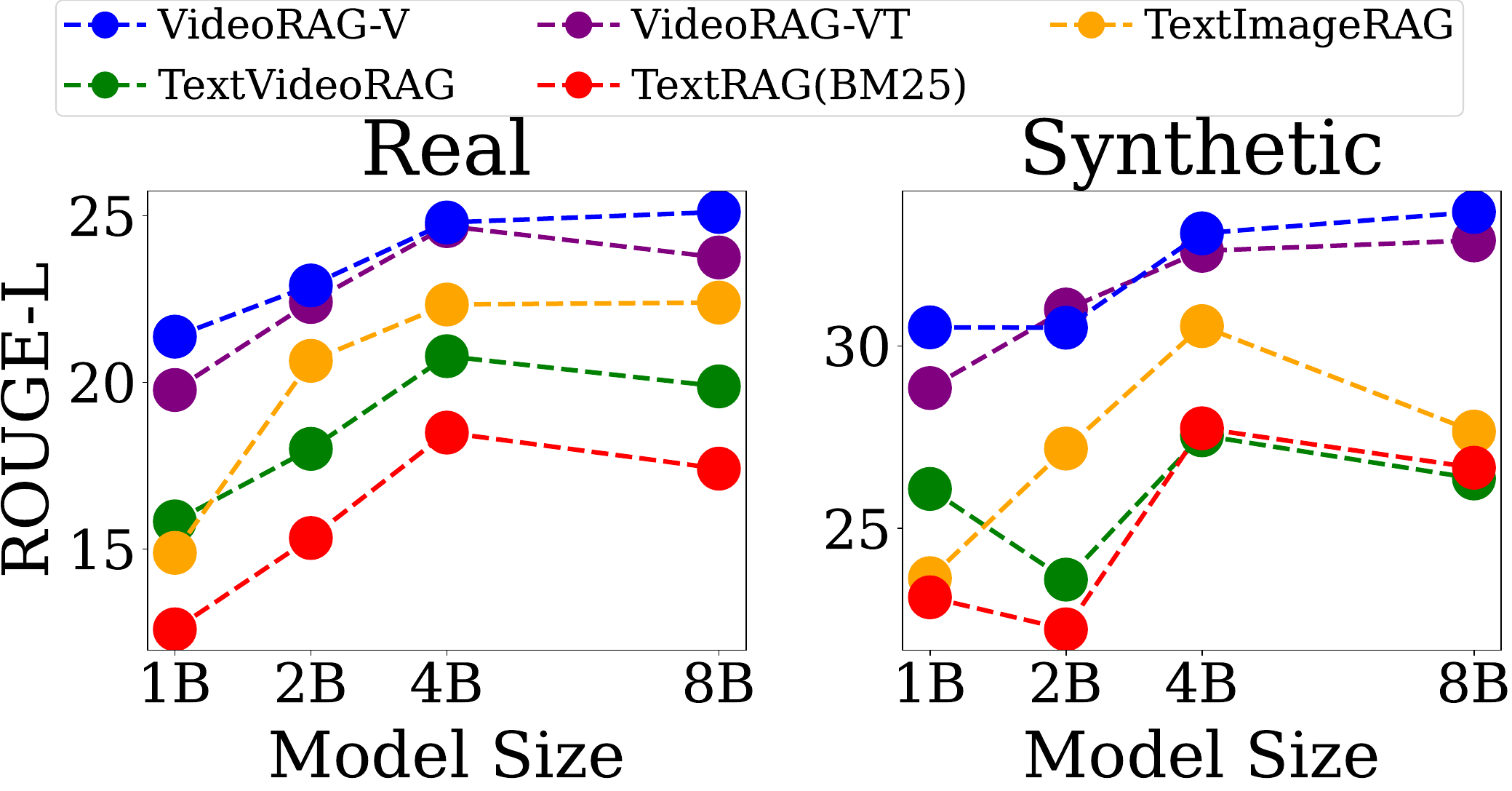}
        \vspace{-0.11in}
        \caption{\small Results of varying InternVL sizes.}
        \label{fig:modelsize}
        \vspace{-0.125in}
    \end{minipage}
    \hfill
    \begin{minipage}{0.63\linewidth}
        \centering
        \includegraphics[width=0.975\linewidth]{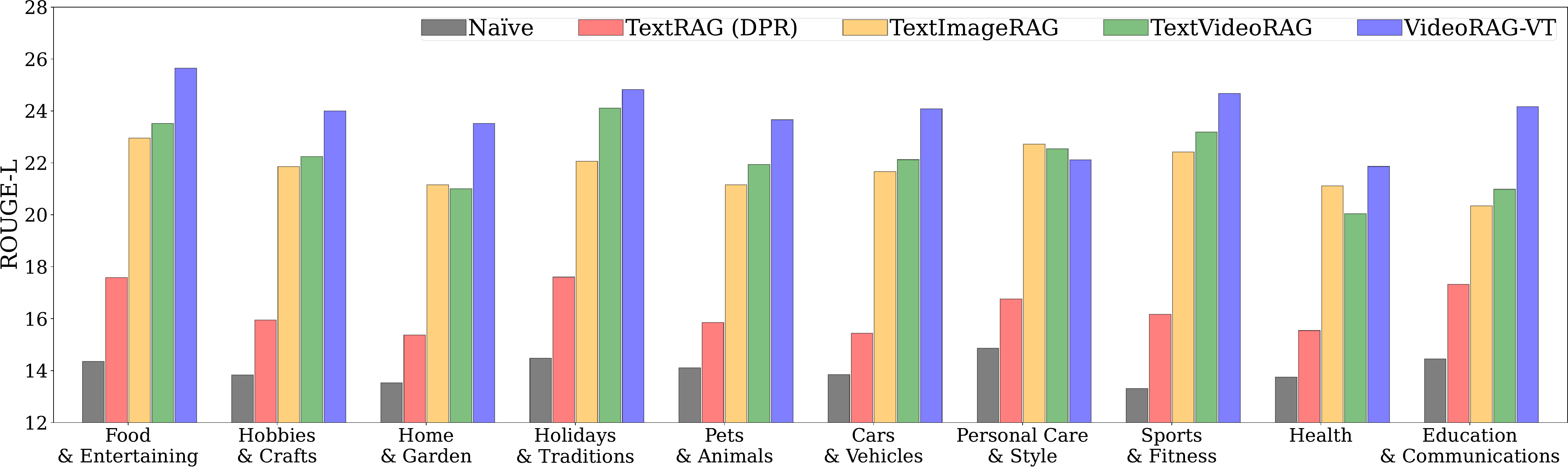}
        \vspace{-0.1in}
        \caption{\small Breakdown performance of different models across 10 categories.}
        \vspace{-0.075in}
    \label{fig:category}
    \end{minipage}
\end{figure*}

\paragraph{Impact of Video Retrieval}
We hypothesize that the quality of the retrieved videos is a critical factor in the success of RAG, as it can directly influence the subsequent answer generation process. To confirm this, we compare the performance of our \textsc{\ours} with retrieved videos against the one with the Oracle setting (which represents an ideal scenario with perfectly relevant video retrieval). Then, Table~\ref{tab:main} shows that the Oracle setting achieves the highest performance, highlighting the potential for further improvements through advancements in video retrieval mechanisms within our VideoRAG.

\paragraph{Efficacy of Textual and Visual Features}
When performing video retrieval, it is questionable how much different modalities, such as textual, visual, or a combination of both, contribute to video representations, and we report results with varying modalities in Table~\ref{tab:feat_ensemble}. We observe that textual features consistently outperform visual features, likely due to their stronger semantic alignment with textual user queries. To further examine this, we visualize the embeddings of textual and visual features of video content as well as queries over the latent space in Figure~\ref{fig:pca}, and it clearly reveals closer proximity between textual query embeddings and textual video representations compared to visual video representations. This is likely due to a modality gap that visual features exhibit relative to text-based queries, resulting in suboptimal retrieval performance. Nevertheless, combining textual and visual features achieves the highest performance, demonstrating the complementary nature of those two modalities in video representations for retrieval.
 
\begin{table}[t]
\centering
\caption{Performance comparison of uniform sampling and our frame selection approach on retrieval and generation tasks.}
\vspace{-0.1in}
\label{tab:frameselect}
\renewcommand{\arraystretch}{1.0}
\setlength{\tabcolsep}{10pt}
\resizebox{\linewidth}{!}{%
\begin{tabular}{p{2mm}lccc}
    \toprule
    & \textbf{Retrieval} & \textbf{R@1} & \textbf{R@5} & \textbf{R@10} \\
    \midrule
    \midrule
    \multirow{2}{*}{\rotatebox[origin=c]{90}{\small \textbf{Visual}}}
    & \textbf{Uniform} & 0.054 & 0.193 & 0.288 \\
    & \textbf{Adaptive (Ours)}    & \textbf{0.079} & \textbf{0.249} & \textbf{0.367} \\
    \noalign{\vskip 0.25ex}\cdashline{1-5}\noalign{\vskip 0.75ex} 
    
    \multirow{2}{*}{\rotatebox[origin=c]{90}{\small \textbf{Ens.}}}
    & \textbf{Uniform} & 0.097 & 0.305 & 0.448 \\
    &\textbf{Adaptive (Ours)}    & \textbf{0.118} & \textbf{0.324} & \textbf{0.453} \\
    \bottomrule
    \toprule
    & \textbf{Generation} & \textbf{ROUGE-L} & \textbf{BLEU-4} & \textbf{BERTScore} \\
    \midrule
    \midrule
    & \textbf{Uniform} & 21.04 & 3.249 & 86.07 \\
    &\textbf{Adaptive (Ours)}    & \textbf{23.24} & \textbf{3.963} & \textbf{87.13} \\
    \bottomrule
\end{tabular}
}
\vspace{-0.1in}
\end{table}

\paragraph{Analysis on Feature Ensemble}
To better understand the contribution of textual and visual features in video retrieval, we analyze how varying their combination ratio ($\alpha$) impacts performance across different metrics. As shown in Figure~\ref{fig:alpha}, the optimal ratio for balancing textual and visual features is around 0.5 to 0.7 (with marginal variations depending on metrics). These results further highlight the complementary contributions of textual and visual features in video representations for retrieval, while a slight emphasis on textual features might be preferable due to the modality gap (Figure~\ref{fig:pca}). 

\paragraph{Effectiveness of Frame Selection}
We analyze the efficacy of our adaptive frame selection, comparing it against uniform sampling in retrieval and generation. Table~\ref{tab:frameselect} shows that our strategy outperforms uniform sampling in both tasks, demonstrating its ability to select more useful frames. Qualitative results in Table~\ref{tab:case_ret} for retrieval and Tables~\ref{tab:case_gen_cut} and~\ref{tab:case_gen_coconut} for generation also highlight the advantage of frame selection over uniform sampling (whose frames are often redundant or less relevant).

\paragraph{Analysis with Varying Model Sizes}
To see if VideoRAG can be instantiated with varying sizes of LVLMs, we report its performance with different InternVL2.5 sizes in Figure~\ref{fig:modelsize}. Then, the performance of \textsc{VideoRAG} improves as the model size increases (thanks to the superior capability of video understanding in larger models), demonstrating the scalability of our VideoRAG and further suggesting its potential benefit with even larger LVLMs.

\begin{table}[t!]
\caption{\small Ablation studies with different modalities. For \textsc{TextRAG}, we use BM25 to retrieve textual documents.}
\vspace{-0.1in}
\small
\centering
\renewcommand{\arraystretch}{1.2}
\setlength{\tabcolsep}{5pt} 
\newcommand*{\mline}[1]{%
    \begingroup
        \renewcommand*{\arraystretch}{1.25}%
        \begin{tabular}[c]{@{}>{\raggedright\arraybackslash}p{3.2cm}@{}}#1\end{tabular}%
    \endgroup
}
\resizebox{\linewidth}{!}{%
    \begin{tabular}{lccccc} 
        \toprule
        \textbf{Methods} & \textbf{Document} & \textbf{Video} & \textbf{Subtitle} & \textbf{ROUGE-L} & \textbf{G-Eval} \\
        \midrule
        \midrule
        \textbf{\textsc{Naïve}} & $\bigtimes$ & $\bigtimes$ & $\bigtimes$ & 14.08 & 1.579 \\
        \textbf{\textsc{TextRAG (BM25)}} & $\bigcirc$ & $\bigtimes$ & $\bigtimes$ & 17.22 & 1.633 \\
        \textbf{\textsc{TextVideoRAG}} & $\bigtimes$ & $\bigtimes$ & $\bigcirc$ & 22.44  & 2.001  \\
        \textbf{\textsc{\ours-VT}} & $\bigtimes$ & $\bigcirc$ & $\bigcirc$ & \textbf{25.23} & \textbf{2.104} \\
        \textbf{\textsc{\ours-VT + TextRAG}} & $\bigcirc$ & $\bigcirc$ & $\bigcirc$ & 24.35 & 2.048 \\
        \bottomrule
    \end{tabular}%
}
\label{tab:combination}
\vspace{-0.1in}
\end{table}

\paragraph{Category-Wise Performance Analysis}
To evaluate the robustness of VideoRAG across diverse query types, we break down the performance on 10 categories (annotated in WikiHow). As shown in Figure~\ref{fig:category}, \textsc{VideoRAG-VT} outperforms all baselines across all categories (except for one), which highlights its ability to handle a variety of queries. Also, \textsc{VideoRAG-VT} shows notable performance gain in a \textit{Food \& Entertaining} category, and this is particularly reasonable given that questions in this category often benefit from visual details; for example, the query: \textit{``How to make a healthy spinach and garlic dish''} requires ingredient preparation or cooking techniques, which are not effectively conveyed through text alone. Thus, the results in this category reaffirm the importance of leveraging video content as external knowledge for RAG.

\paragraph{Ablation Studies}
To analyze how performance varies with different knowledge sources, we conduct ablation studies and present results in Table~\ref{tab:combination}. From this, we then observe that, while incorporating external knowledge (whether from textual encyclopedic sources or video corpus) consistently improves performance over the \textsc{Naïve} baseline, the approach that jointly uses videos with general textual documents achieves slightly degraded performance. This suggests that textual content (retrieved from the encyclopedic knowledge base) may introduce redundant or irrelevant details, which may overlap with or contradict the information provided by video content, leading to a diminishing effectiveness of the VideoRAG framework.

\begin{table*}[ht]
\caption{\small Case study comparing \textsc{Naïve} and \textsc{VideoRAG-V} approaches.}
\vspace{-0.1in}
\centering
\footnotesize 
\resizebox{\textwidth}{!}{
\renewcommand{\arraystretch}{0.5}
\begin{tabular}{l c}
\toprule
\multicolumn{2}{c}{\textbf{\textsc{\large Naïve}}} \\ 
\midrule
\textbf{Question} & \multicolumn{1}{p{.99\textwidth}}{Explain how to bake cookies on your car dashboard.} \\
\noalign{\vskip 0.25ex}\cdashline{1-2}\noalign{\vskip 0.75ex}
\textbf{Generated Answer} & \multicolumn{1}{p{.99\textwidth}}{I'm sorry, but it's not possible to bake cookies on your car dashboard. The dashboard is not designed for cooking and it's not safe to use it as a heat source. Additionally, the fumes from baking could be harmful to you and the car's interior.\newline\newline Baking requires a controlled environment with a consistent temperature, which is difficult to achieve in a car. Moreover, the dashboard is made of materials that can easily catch fire or release toxic fumes when exposed to high temperatures.\newline\newline If you want to bake cookies, it's best to do so in a safe and controlled environment, such as a kitchen or an outdoor grill. Just remember to always follow proper safety precautions and guidelines when cooking.} \\
\midrule
\multicolumn{2}{c}{\textbf{\large \textsc{\ours-V}}} \\
\midrule
\textbf{Question} & \multicolumn{1}{p{.99\textwidth}}{Explain how to bake cookies on your car dashboard.} \\
\noalign{\vskip 0.25ex}\cdashline{1-2}\noalign{\vskip 0.75ex}
\textbf{Retrieved Video} & \parbox[t]{.99\textwidth}{\vspace{-3pt}\includegraphics[width=.8\textwidth]{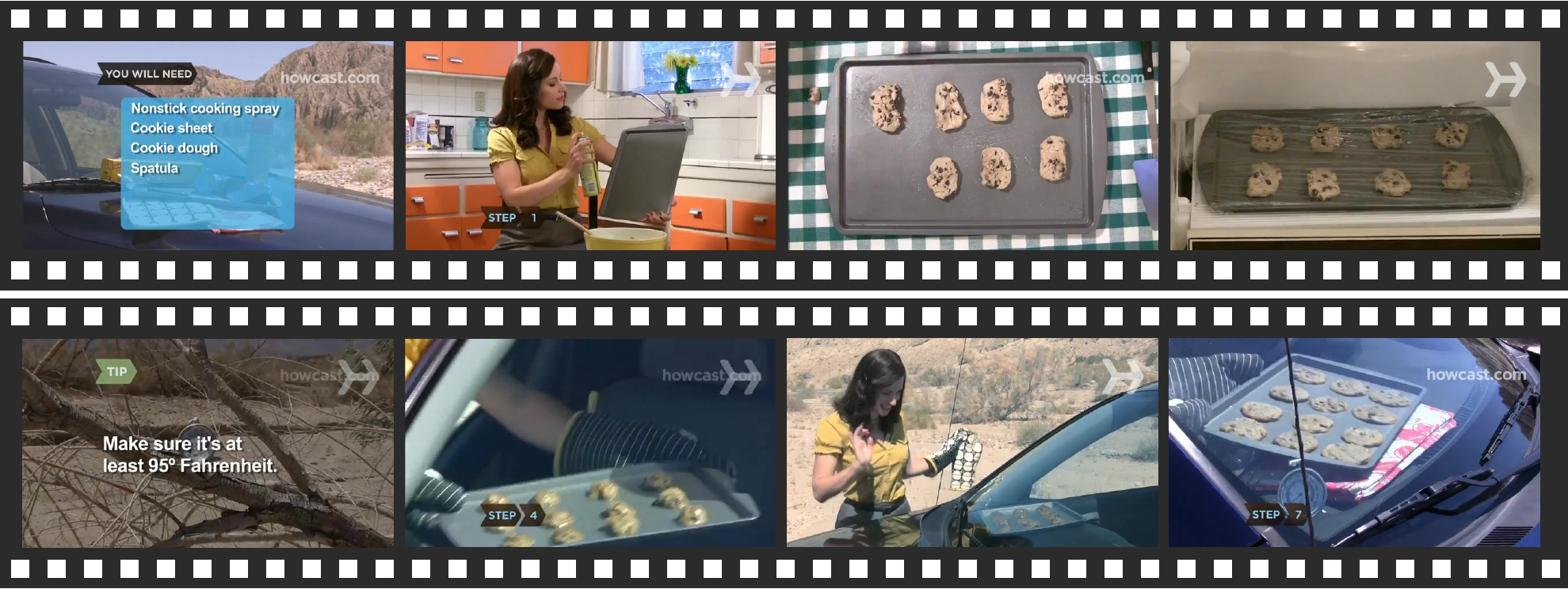}\vspace{5pt}} \\
\noalign{\vskip 0.25ex}\cdashline{1-2}\noalign{\vskip 0.75ex}
\textbf{Generated Answer} & \multicolumn{1}{p{.99\textwidth}}{To bake cookies on your car dashboard, you'll need a nonstick cooking spray, a cookie sheet, cookie dough, a spatula, and a thermometer. First, spray the cookie sheet with nonstick cooking spray. Then, place the cookie dough onto the sprayed sheet. Cover it with plastic wrap and let it rest. Next, place the cookie sheet in the car's dashboard, ensuring it's at least 95 degrees Fahrenheit. Set the car in a sunny spot to bake the cookies. After baking, use a spatula to remove the cookies from the dashboard.} \\
\midrule
\midrule
\textbf{Ground Truth} & \multicolumn{1}{p{.99\textwidth}}{To bake cookies on your car dashboard, you'll need to be parked in a sunny spot on a day when it's at least 95 degrees Fahrenheit out. First, spread cookie dough out on a baking sheet lined with parchment paper. Then, set the baking sheet on your car dashboard, and shut all of the doors. Finally, check on the cookies every 15-30 minutes until the edges of the cookies are firm and you're able to slide them off the parchment paper.} \\
\bottomrule

\end{tabular}
}
\label{tab:qualitative2}
\vspace{-0.05in}
\end{table*}

\begin{table}[t]
\centering
\caption{Human evaluation results. The results are evaluated with the subset of WikiHowQA over the HowTo100M corpus.}
\vspace{-0.1in}
\label{tab:human_eval}
\renewcommand{\arraystretch}{0.25}
\setlength{\tabcolsep}{17pt}
\resizebox{\linewidth}{!}{%
\begin{tabular}{lcc}
    \toprule
    \textbf{Methods} & \textbf{Human} & \textbf{G-Eval} \\
    \midrule
    \midrule
    \textbf{\textsc{Naïve}} & 1.833 & 1.684\\
    \textbf{\textsc{TextRAG (DPR)}} & 1.867 & 1.747 \\
    \textbf{\textsc{TextImageRAG}} & 2.447 & 2.203 \\
    \textbf{\textsc{TextVideoRAG}} & 3.130 & 2.279 \\
    \textbf{\textsc{VideoRAG-VT}} & \textbf{4.043} & \textbf{3.689} \\
    \bottomrule
\end{tabular}
}
\vspace{-0.125in}
\end{table}

\paragraph{Human Evaluation}
To complete automatic metrics, we conduct a human evaluation. Specifically, we recruit 12 evaluators and split (randomly sampled) 50 queries into two sets of 25, assigning each participant to assess one (including responses from four baselines and our model) with a 5-point Likert scale. The results, presented in Table~\ref{tab:human_eval}, show that our VideoRAG achieves the highest performance in human evaluation. Further, to validate the quality and reliability of human evaluation, we measure an inter-annotator agreement among annotators who evaluate the same subset, by using Spearman's correlation coefficient between the ranked scores of different annotators. Then, we obtain a coefficient of 0.632, confirming the high reliability of our assessments. Similarly, we measure the agreement between human- and model-based (G-Eval) evaluations and obtain a coefficient of 0.588, indicating that G-Eval is a reasonable proxy for judgment. 

\paragraph{Case Study}
Lastly, we provide a case-study example, with the query: ``\textit{Explain how to bake cookies on your car dashboard}''. As shown in Table~\ref{tab:qualitative2}, the \textsc{Naïve} baseline, relying solely on its parametric knowledge, generates a generic response highlighting the impracticality and safety concerns of such a method, failing to provide the step-by-step instructions necessary to address the query. This example indicates the limitation of parametric knowledge that is inadequate, especially when specific and uncommon information is required. In contrast, \textsc{VideoRAG-V} retrieves the relevant video that illustrates the process of baking cookies on a car dashboard, and, by leveraging this, it successfully generates a response similar to the ground truth. This highlights how VideoRAG utilizes external video content to produce more precise, contextually rich, and actionable answers. We provide an additional example in Table~\ref{tab:qualitative1} of Appendix~\ref{appendix:qualitative}.

\section{Related Work}

\paragraph{Retrieval-Augmented Generation}
RAG is a strategy that combines retrieval and generation processes to produce accurate answers by grounding them in external knowledge~\cite{ralm, rag_survey1}. To be specific, during the retrieval step, documents (relevant to queries) are selected from a large corpus by calculating their similarity to the query, which can be done with retrievers~\cite{bm25, Jones2004tfidf, dpr, contriever}. In the generation step, these retrieved documents serve as input for generating answers that are rooted in the provided information~\cite{flare, self-rag, hwang-etal-2024-dslr, uar}, with some advancements using iterative retrieval-generation cycles~\cite{ircot} or adapting different RAG strategies based on query complexity~\cite{adaptive-rag}. However, despite the fact that much of the real-world knowledge is inherently multimodal in nature~\cite{Lee2024UnifiedMI, daqu, ColPali}, the majority of RAG studies have focused on the textual modality, with little effort on incorporating images, leaving a significant gap in leveraging the full spectrum of available knowledge for the holistic operation of RAG. 

\paragraph{Multimodal RAG}
There has been growing interest in expanding RAG to incorporate multimodal information (beyond text), such as images~\cite{MuRAG, DBLP:conf/emnlp/LinB22, DBLP:journals/corr/abs-2410-21943, visRAG}, code~\cite{DBLP:conf/nlpcc/GuoLJLZLLYBGC24}, tables~\cite{trag, DBLP:journals/corr/abs-2408-14717}, and audio~\cite{DBLP:conf/icassp/YuanLLHP024}. However, unlike them, videos offer a unique and orthogonal advantage for RAG, as they encapsulate temporal dynamics, spatial details, and multimodal cues in ways unmatched by other modalities. Inspired by this fact, very recent studies have started exploring the usage of video content within RAG pipelines; however, existing approaches leverage it in a suboptimal way. To be specific, some focus on extracting query-relevant frames from the preselected video and generating answers based on them, which, while useful in controlled scenarios, limits their real-world applicability in open-domain settings~\cite{videorag,drvideo}. Also, some other studies attempt to sidestep the complexity of handling video data by converting it into textual representations (such as subtitles or captions); however, while directly applicable to existing text-based RAG frameworks, they sacrifice the multimodal richness embedded within videos (such as temporal dynamics and spatial patterns)~\cite{iRAG,OmAgent,drvideo}. To address these, we propose VideoRAG which is capable of dynamically retrieving and holistically utilizing video content in RAG, powered by LVLMs discussed next.

\paragraph{Large Video Language Models}
Building on the remarkable success of LLMs~\cite{GPT-4, Gemini, llama3, DBLP:conf/naacl/ChoCHSJLSPK25,song2025doesrationalequalitymatter}, there has been a growing interest in extending them to encompass diverse modalities, such as images~\cite{MM-Embed, vlm_intro, vlm_survey} and code~\cite{DeepSeek, Qwen2.5-Coder}. Also, this expansion has recently extended to another modality called video, leading to the emergence of LVLMs that are capable of directly processing video content. They excel in solving traditionally challenging (yet straightforward) tasks, such as object or action detection, and their capabilities have rapidly advanced to tackle more challenging tasks, such as analyzing spatio-temporal dynamics to predict event sequences, inferring causal relationships, and generating context-aware descriptions of intricate scenarios~\cite{lvlm_survey, DBLP:conf/eccv/WangYWNH24, Video-ChatGPT, DBLP:conf/cvpr/ZhangLYWLL0W24, DBLP:conf/cvpr/0004LJJCSSL24,DBLP:conf/cvpr/Wang0LHYWJ24, DBLP:conf/naacl/HwangCLP25}, even in zero-shot settings~\cite{VideoLLM-online, Kim2024VideoICLCI}. However, their potential has yet to be explored in the context of RAG; thus, in this work, we aim to bridge this gap with VideoRAG.

\section{Conclusion}
\vspace{-0.025in}
We presented VideoRAG, a framework that expands the current landscape of RAG by leveraging a video corpus as the external knowledge source. Specifically, unlike existing works that use the textual representations of videos or assume the existence of query-relevant videos without retrieval, the proposed VideoRAG retrieves videos based on their relevance to queries but also integrates their multimodal richness (including visual and textual elements) into the RAG pipeline, with adaptive frame selection to leverage only the most informative subset of full frames for effectiveness and efficiency. Also, through comprehensive analyses, we demonstrated how the inclusion of visual or textual features, or a combination of both, improves retrieval and generation performance, and, inspired by the critical role of textual features (for retrieval quality) but their absence in some videos, we presented a simple yet effective mitigator that uses automatic speech recognition to generate textual transcripts. Overall, experimental results validated the superiority of our VideoRAG over existing RAG methods, and we believe it makes a significant step toward holistic RAG systems that can utilize videos.

\section*{Limitations}
It is worth noting that our VideoRAG is one of the first works that operationalizes the full pipeline of RAG over the video corpus, including dynamic retrieval of query-relevant videos and answer generation grounded in them, and to evaluate this operation, the set of triples for query, relevant videos, and ground-truth answers is required. However, we discover that such datasets are currently limited, and to tackle this issue, we not only construct the dataset by associating the WikiHowQA dataset (providing pairs of query and answers) with the HowTo100M dataset (providing pairs of query and videos), but also automatically collect the synthetic dataset. While this process enables a comprehensive evaluation, it would be also valuable as a future work to develop and release the benchmark dataset, to greatly facilitate research on RAG over videos. Additionally, the proposed frame selection strategy greatly improves the efficiency of video processing for retrieval and generation (as it narrows down the entire frames for the given video into their small subset) as well as their effectiveness, and it would be interesting future work to further improve the efficacy of our initial foray (VideoRAG) by maximizing its effectiveness and efficiency further.

\section*{Ethics Statement}
Recall that our proposed VideoRAG is designed to offer answers to user queries by retrieving query-relevant videos from a large video corpus, which helps enhance response quality. Yet, the retrieval process inherently depends on the corpus, and if it includes biased, harmful, or otherwise problematic examples, it may lead to generating responses that reflect those issues. In addition, since the generation process is powered by LVLMs, which are trained on vast multimodal datasets, their responses may inherit and amplify biases present in their training data. Therefore, we recommend practitioners to carefully evaluate those potential risks and consider mitigating them with some strategies, for example, bias detection and filtering~\cite{DBLP:conf/acl/ShinSLJP24, DBLP:conf/nips/MiaoZY0GD24, DBLP:conf/emnlp/LeeSSCHP24, DBLP:journals/corr/abs-2503-09669}.

\section*{Acknowledgements}
This work was supported by 
the National Research Foundation of Korea (NRF) grant funded by the Korea government (MSIT) (No. RS-2023-00256259),
the Institute for Information \& communications Technology Planning \& Evaluation (IITP) grant funded by the Korea government (MSIT) (RS-2019-II190075, Artificial Intelligence Graduate School Program (KAIST)),
the Institute of Information \& communications Technology Planning \& Evaluation (IITP) grant funded by the Korea government (MSIT) (No.RS-2022-II220713, Meta-learning Applicable to Real-world Problems),
the Artificial intelligence industrial convergence cluster development project funded by the Ministry of Science and ICT (MSIT, Korea) \& Gwangju Metropolitan City,
the grant of the Korea Machine Learning Ledger Orchestration for Drug Discovery Project (K-MELLODDY), funded by the Ministry of Health \& Welfare and Ministry of Science and ICT, Republic of Korea (grant number: RS-2024-12345678)
the Center for Applied Research in Artificial Intelligence (CARAI) grant funded by DAPA and ADD (UD230017TD),
and the Institute of Information \& Communications Technology Planning \& Evaluation (IITP) with a grant funded by the Ministry of Science and ICT (MSIT) of the Republic of Korea in connection with the Global AI Frontier Lab International Collaborative Research. (No. RS-2024-00469482 \& RS-2024-00509279)

\bibliography{custom}

\clearpage
\appendix

\section{Additional Implementation Details}
\label{sec:appendix}

\subsection{Details on Choice of LVLMs for~Retrieval~and~Generation}
\label{sec:appen_model_choice}
It is worth noting that there exist various LVLMs available for use, each with different merits depending on the task requirements: for retrieval, precise alignment between textual and video features (obtained from their specialized encoders) is essential to ensure that the retrieved videos are contextually relevant to the query, meanwhile, generation benefits from LVLMs with advanced capabilities for accurately formulating responses and grounding them in the retrieved content. To achieve this, for retrieval, we use InternVideo2~\cite{internvideo2} since it is explicitly trained to align semantics between videos and their textual descriptions. Specifically, we use its video and text encoders to extract embeddings for videos and text, respectively. On the other hand, for video-augmented answer generation, we use LLaVA-Video, InternVL 2.5, and Qwen-2.5-VL~\cite{llavanext-video, internvl2.5, Qwen2.5-VL}, which are known for achieving state-of-the-art performance on video understanding and relevant tasks. Finally, for generation, we retrieve and use one video, as we observe that there are not many differences in generation performance with different video quantities, while increasing the number of augmented videos substantially increases the computational costs.

\subsection{Details on Synthetic Data Generation}
\label{sec:appen_syn_data_gen}
To more thoroughly evaluate the effectiveness of our VideoRAG framework, we further automatically generate question-answer pairs grounded in individual videos via prompting of LVLMs (in addition to utilizing the real-world benchmark dataset). Specifically, since our objective is to retrieve query-relevant videos from a large corpus, the generated questions should not be overly specific to a single video; for example, frame-specific questions like \textit{``In this video, what is the color of the balloon that the girl popped?''}. Instead, they should be formulated in a more general manner to facilitate the retrieval of multiple relevant videos, such as \textit{``After mashing the ingredients for a homemade prison beer, what is the next crucial step?''}. To achieve this, we construct a structured prompt for the LLM, providing context about RAG and outlining key principles for question generation, such as instructing the model to create three diverse, well-formed question-answer pairs that leverage the video content without being overly specific and suitable for the RAG framework. We provide the prompt used to elicit the generation of question-answer pairs in Table~\ref{tab:prompt:synthetic}. Also, we use the state-of-the-art GPT-4o as the LVLM for the synthetic data creation.

\subsection{Additional Details on Frame Selection}
\label{sec:appen_frame_selection}
We discuss how we instantiate the scoring function $f$ (whose goal is to assign a score to the subset of frames) for retrieval and generation, and how we train it with the dataset automatically collected from the training dataset, as follows:

\paragraph{Retrieval} 
In retrieval, to efficiently handle a large number of videos within the corpus, we set the number of frames extracted from the frame selection process as four. Specifically, for each video, we first sample its frames at 1 fps and extract their features with CLIP. Also, as discussed in Section~\ref{sec:frame_selection}, to eliminate redundancy and ultimately reduce the frame sampling space, we apply $k$-means++ clustering and extract 8 candidate frames, leading to the smaller sampling space of $\comb{8}{4}$. The objective of $f$ then becomes scoring the set of 4 frames, and we design this by obtaining the representations for those 4 frames from CLIP and passing their concatenated representations through 3-layer MLPs. Also, this MLP network is trained with the automatically collected labels to obtain the most representative frames for a certain video that lead to the retrieval success, where the retrieval success is decided by the high similarity between the selected frames of a certain video and its associated query. In other words, given the pair of the query and its relevant video, we sample multiple sets of 4 frames, and measure their similarities with the given query, so that we label the top 3 combinations with the highest similarities as True and the bottom 3 combinations as False. Then, the network is optimized via cross-entropy loss based on these labels.

\paragraph{Generation} 
Similar to how we select frames for retrieval, in generation, we aim to select 32 frames from 64 candidate frames (obtained via $k$-means++ clustering). Notably, the number of frames is larger than the retrieval as generation benefits more from a comprehensive understanding of the video content to improve response accuracy. Also, among the resulting $\comb{64}{32}$ possible combinations, we randomly sample 40 subsets as the space of $\comb{64}{32}$ is still very large. For the scoring function $f$, we design this by obtaining representations of sampled frames as well as the query (to consider their relevance with it) from 3-layer MLPs on top of CLIP, and then computing the dot product between the averaged frame representation and the query representation. Also, we automatically collect the training dataset by labeling the top 3 combinations with the highest ROUGE-L scores as True and the bottom 3 with the lowest scores as False, according to their ROUGE-L score and with the LLaVA-Video (7B) as the LVLM for generation.

\begin{table}[t!]
\caption{\small Generation results using a different set of videos, such as Random that randomly samples videos, Retrieved that selects videos according to their relevance with queries, and Oracle that uses the ground truth videos annotated in data.}
\vspace{-0.1in}
\small
\centering
\renewcommand{\arraystretch}{1.0}
\setlength{\tabcolsep}{10pt}
\newcommand{\oracle}[1]{\textcolor{gray}{#1}}
\resizebox{\linewidth}{!}{%
    \begin{tabular}{lcccc}
    \toprule
    \textbf{Video Set} & \textbf{ROUGE-L} & \textbf{BLEU-4} & \textbf{BERTScore} \\
    \midrule
    \midrule
    \textbf{Random} & 24.29 & 4.996 & 87.83 \\
    \textbf{Retrieved} & \textbf{25.42} & \textbf{5.375} & \textbf{88.12} \\
    \textbf{Oracle} & 26.19 & 5.480 & 88.41 \\ \bottomrule
    \end{tabular}
}
\vspace{-0.05in}
\label{tab:video_impact}
\end{table}

\section{Impact of Videos on Answer Quality} 
As an auxiliary analysis, we compare the performance of our VideoRAG augmented with different videos, including randomly selected videos and retrieved videos (relevant to queries). As shown in Table~\ref{tab:video_impact}, incorporating query-relevant videos significantly improves the quality of answers compared to randomly selected videos, demonstrating the importance of retrieval quality. Furthermore, the Oracle setting, which represents an ideal scenario with perfectly relevant video retrieval, achieves the highest performance, highlighting the potential for further improvements through advancements in video retrieval mechanisms within our VideoRAG.

\section{Effectiveness of Frame Reduction}
To further validate our choice of $k$-means++ clustering when reducing the full set of frames to a smaller subset to obtain a diverse yet representative subset of $k$ frames, we perform comparative experiments using alternative frame reduction operations, including random sampling (which randomly samples multiple subsets of $n$ frames from the entire video) and uniform sampling (which selects $k$ frames and then samples $n$ frames among $k$, similar to ours). As shown in Table~\ref{tab:kmeans}, we observe that $k$-means consistently outperforms these alternatives, suggesting that clustering-based reduction provides a better initialization for the final frame selection. Nonetheless, VideoRAG is flexible, allowing anyone to replace the current frame reduction operation of $k$-means with others, which would be interesting for future work.

\begin{table}[t]
\centering
\caption{Comparison of video retrieval performance using three different frame reduction methods on the WikiHowQA and SyntheticQA datasets. The retrieval performance is measured by R@1.}
\vspace{-0.1in}
\label{tab:kmeans}
\renewcommand{\arraystretch}{0.875}
\setlength{\tabcolsep}{10pt}
\resizebox{\linewidth}{!}{%

    \begin{tabular}{lcc}
    \toprule
    \textbf{Method} & \textbf{WikiHowQA} & \textbf{SyntheticQA} \\
    \midrule
    \midrule
    \textbf{Random} & 0.101 & 0.103 \\
    \textbf{Uniform} & 0.099 & 0.094 \\
    \textbf{Clustering (Ours)} & \textbf{0.118} & \textbf{0.122} \\ 
    \bottomrule
    \end{tabular}

}
\vspace{-0.05in}
\end{table}

\section{Qualitative Results}
\label{appendix:qualitative}
We now qualitatively analyze the effectiveness of VideoRAG through a case study, in addition to the example shown in Table~\ref{tab:qualitative2}.
As shown in Table~\ref{tab:qualitative1}, we observe that external textual knowledge alone can sometimes fall short in providing relevant and actionable information for specific procedural queries, such as ``\textit{Explain how to make a clay rose}''. To be more specific, \textsc{TextRAG (BM25)} retrieves an irrelevant document about a person named Rose, as Wikipedia does not contain specific procedural guidance on this topic, and, consequently, the generated response is misaligned with the query. In contrast, \textsc{VideoRAG-V} retrieves the relevant video demonstrating how to make a clay rose and leverages this visual content to generate a concise and accurate response that closely mirrors the ground truth, from which we clearly confirm the utility of videos for RAG.

\clearpage
\begin{table*}[t!]
\caption{Case study comparing uniform sampling and our frame selection on the retrieval task.}
\vspace{-0.1in}
\centering
\resizebox{\textwidth}{!}{
    \renewcommand{\arraystretch}{0.8}
    \begin{tabular}{c c}
    \toprule
    \textbf{\large Uniform Sampling} & \textbf{\large Adaptive Frame Selection} \\ 
    \midrule
    \multicolumn{2}{c}{\textbf{Make a banana split}} \\
    \includegraphics[width=.5\textwidth]{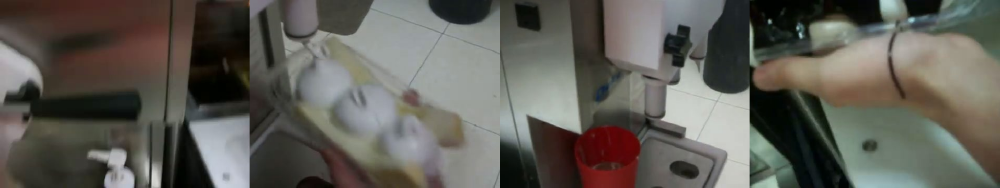} 
    & \includegraphics[width=.5\textwidth]{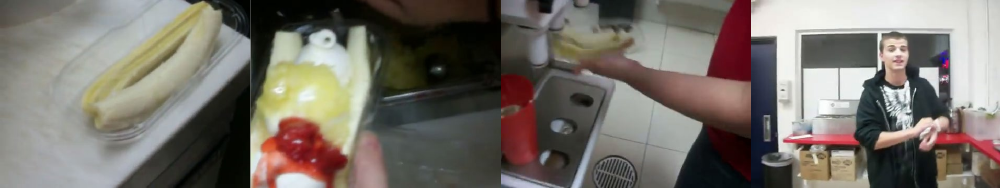} \\
    \noalign{\vskip 0.25ex}\cdashline{1-2}\noalign{\vskip 0.75ex}
    \multicolumn{2}{c}{\textbf{Clean a Nespresso machine}} \\
    \includegraphics[width=.5\textwidth]{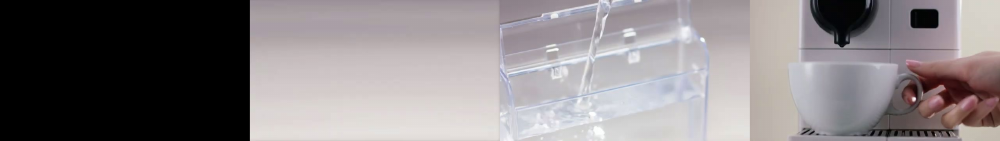} 
    & \includegraphics[width=.5\textwidth]{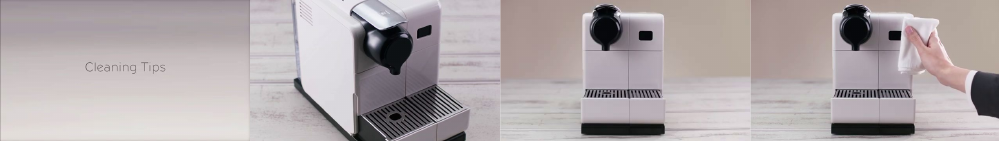} \\
    \noalign{\vskip 0.25ex}\cdashline{1-2}\noalign{\vskip 0.75ex} 
    \multicolumn{2}{c}{\textbf{Cook Italian sausage}} \\
    \includegraphics[width=.5\textwidth]{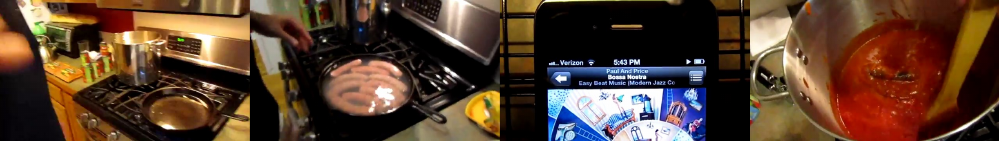} 
    & \includegraphics[width=.5\textwidth]{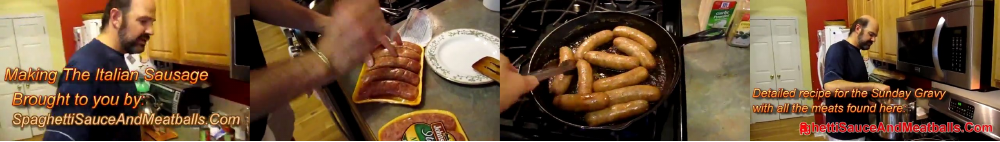} \\
    \noalign{\vskip 0.25ex}\cdashline{1-2}\noalign{\vskip 0.75ex} 
    \multicolumn{2}{c}{\textbf{Clean artificial flowers}} \\
    \includegraphics[width=.5\textwidth]{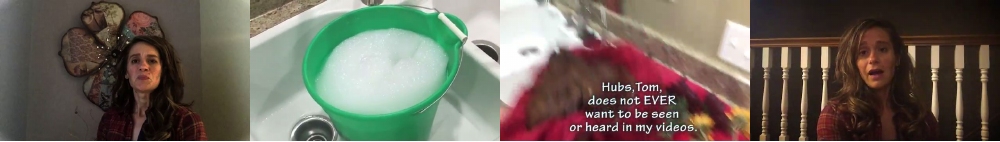} 
    & \includegraphics[width=.5\textwidth]{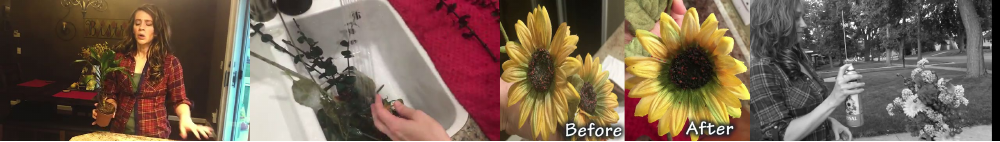} \\
    \bottomrule
    
    \end{tabular}
}
\label{tab:case_ret}
\vspace{-0.1in}
\end{table*}
\begin{table*}[t!]
\caption{Case study comparing uniform sampling and our frame selection on the generation task.}
\vspace{-0.1in}
\scriptsize
\centering
\resizebox{\textwidth}{!}{
\renewcommand{\arraystretch}{0.5}
\begin{tabular}{l c}
\toprule
\multicolumn{2}{c}{\textbf{{\large Uniform Sampling}}} \\ 
\midrule
\textbf{Question} & \multicolumn{1}{p{.9\textwidth}}{Explain how to cut acorn squash.} \\
\noalign{\vskip 0.25ex}\cdashline{1-2}\noalign{\vskip 0.75ex}
\textbf{Sampled Frames} & \parbox[t]{.9\textwidth}{\vspace{-3pt}\includegraphics[width=.8\textwidth]{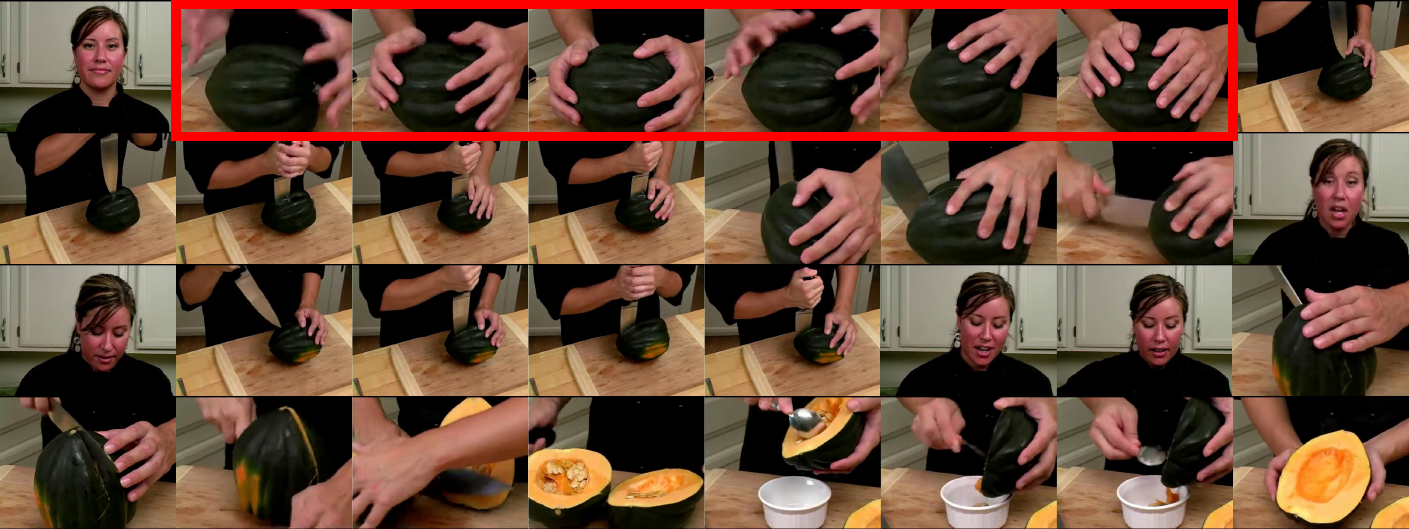}\vspace{5pt}} \\
\noalign{\vskip 0.25ex}\cdashline{1-2}\noalign{\vskip 0.75ex}
\textbf{Generated Answer} & \multicolumn{1}{p{.9\textwidth}}{To cut an acorn squash, first, place it on a cutting board. Use a large knife to make a vertical cut around the middle of the squash, starting from the top and going down to the bottom. Then, make another vertical cut on the opposite side, creating two halves. Remove the seeds and fibrous material using a spoon or a small knife. Finally, scoop out the seeds and any remaining fibrous material into a bowl for later use.} \\
\noalign{\vskip 0.25ex}\cdashline{1-2}\noalign{\vskip 0.75ex}
\textbf{ROUGE-L Score} & \multicolumn{1}{p{.9\textwidth}}{25.50} \\
\midrule
\multicolumn{2}{c}{\textbf{\large Adaptive Frame Selection}} \\
\midrule
\textbf{Question} & \multicolumn{1}{p{.9\textwidth}}{Explain how to cut acorn squash.} \\
\noalign{\vskip 0.25ex}\cdashline{1-2}\noalign{\vskip 0.75ex}
\textbf{Sampled Frames} & \parbox[t]{.9\textwidth}{\vspace{-3pt}\includegraphics[width=.8\textwidth]{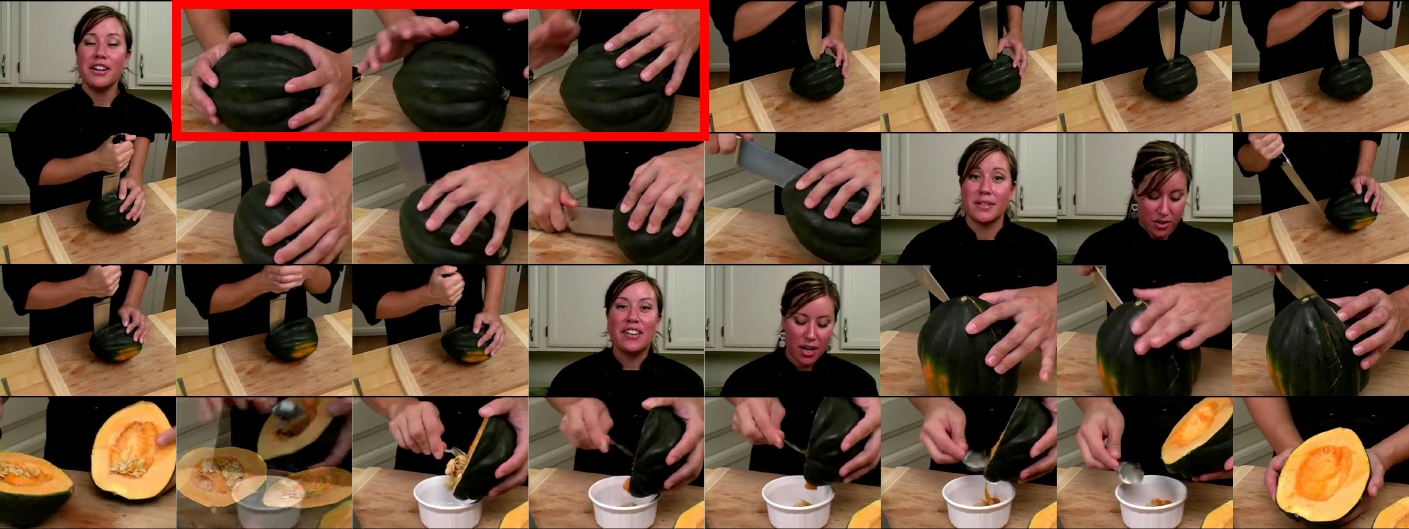}\vspace{5pt}} \\
\noalign{\vskip 0.25ex}\cdashline{1-2}\noalign{\vskip 0.75ex}
\textbf{Generated Answer} & \multicolumn{1}{p{.9\textwidth}}{To cut an acorn squash, first, place it on a cutting board. Use a large knife to cut off the top and bottom ends of the squash. Then, cut the squash in half lengthwise. Remove the seeds and fibrous material from the inside using a spoon or small knife. Finally, scoop out the flesh and place it into a bowl for cooking or serving.} \\
\noalign{\vskip 0.25ex}\cdashline{1-2}\noalign{\vskip 0.75ex}
\textbf{ROUGE-L Score} & \multicolumn{1}{p{.9\textwidth}}{31.11} \\
\midrule
\midrule
\textbf{Ground Truth} & \multicolumn{1}{p{.9\textwidth}}{In order to cut your acorn squash in half, select a sharp, study knife with an 8 to 10 inch blade. Then, place your squash lengthwise on the cutting board and cut 1/2 an inch off the ends of the squash. Next, place your acorn squash sliced-side down so the squash looks like it's standing up. Use a rocking or sawing motion to cut the squash right down the middle.} \\
\bottomrule

\end{tabular}
}
\label{tab:case_gen_cut}
\vspace{-0.1in}
\end{table*}
\begin{table*}[t!]
\caption{Case study comparing uniform sampling and our frame selection on the generation task.}
\vspace{-0.1in}
\scriptsize
\centering
\resizebox{\textwidth}{!}{
\renewcommand{\arraystretch}{0.5}
\begin{tabular}{l c}
\toprule
\multicolumn{2}{c}{\textbf{\large Uniform Sampling}} \\ 
\midrule
\textbf{Question} & \multicolumn{1}{p{.9\textwidth}}{Explain how to make coconut candy.} \\
\noalign{\vskip 0.25ex}\cdashline{1-2}\noalign{\vskip 0.75ex}
\textbf{Sampled Frames} & \parbox[t]{.9\textwidth}{\vspace{-3pt}\includegraphics[width=.8\textwidth]{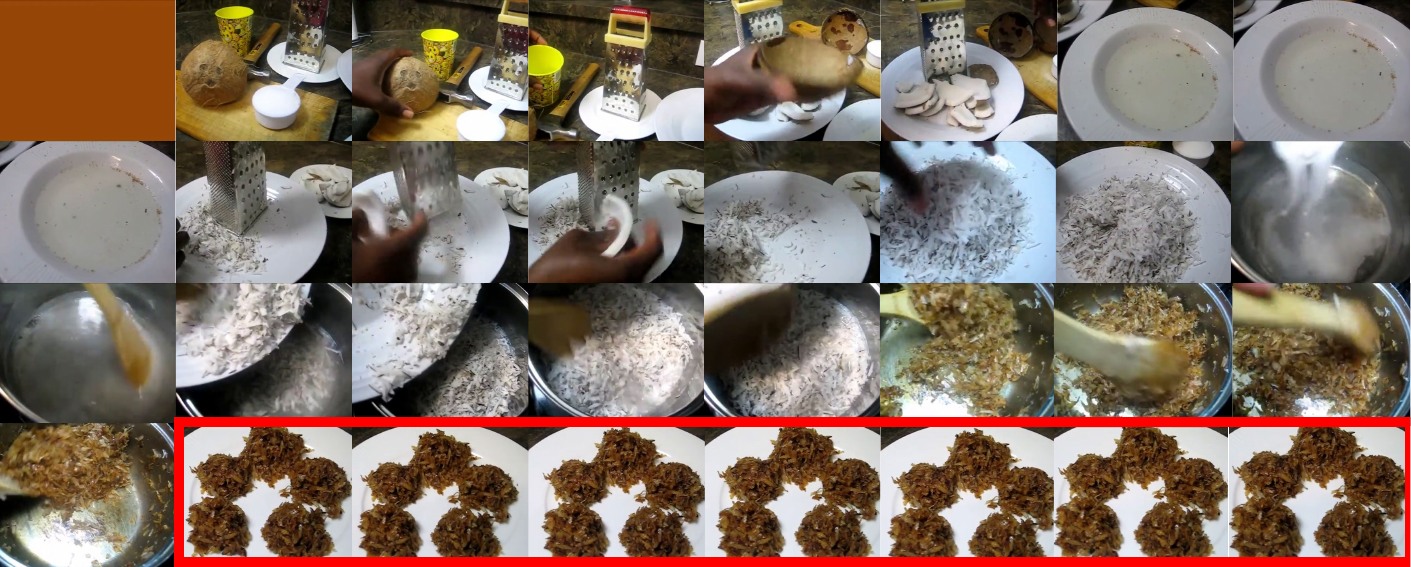}\vspace{5pt}} \\
\noalign{\vskip 0.25ex}\cdashline{1-2}\noalign{\vskip 0.75ex}
\textbf{Generated Answer} & \multicolumn{1}{p{.9\textwidth}}{To make coconut candy, start by grating the coconut and placing it in a pot. Then, add sugar to the pot and stir until the mixture is well combined. Cook the mixture over medium heat until it reaches a caramel-like consistency. Once cooked, remove from heat and let cool slightly. Shape the mixture into small balls and place them on a plate. Let them cool completely before serving.} \\
\noalign{\vskip 0.25ex}\cdashline{1-2}\noalign{\vskip 0.75ex}
\textbf{ROUGE-L Score} & \multicolumn{1}{p{.9\textwidth}}{30.26} \\
\midrule
\multicolumn{2}{c}{\textbf{\large Adaptive Frame Selection}} \\
\midrule
\textbf{Question} & \multicolumn{1}{p{.9\textwidth}}{Explain how to make coconut candy.} \\
\noalign{\vskip 0.25ex}\cdashline{1-2}\noalign{\vskip 0.75ex}
\textbf{Sampled Frames} & \parbox[t]{.9\textwidth}{\vspace{-3pt}\includegraphics[width=.8\textwidth]{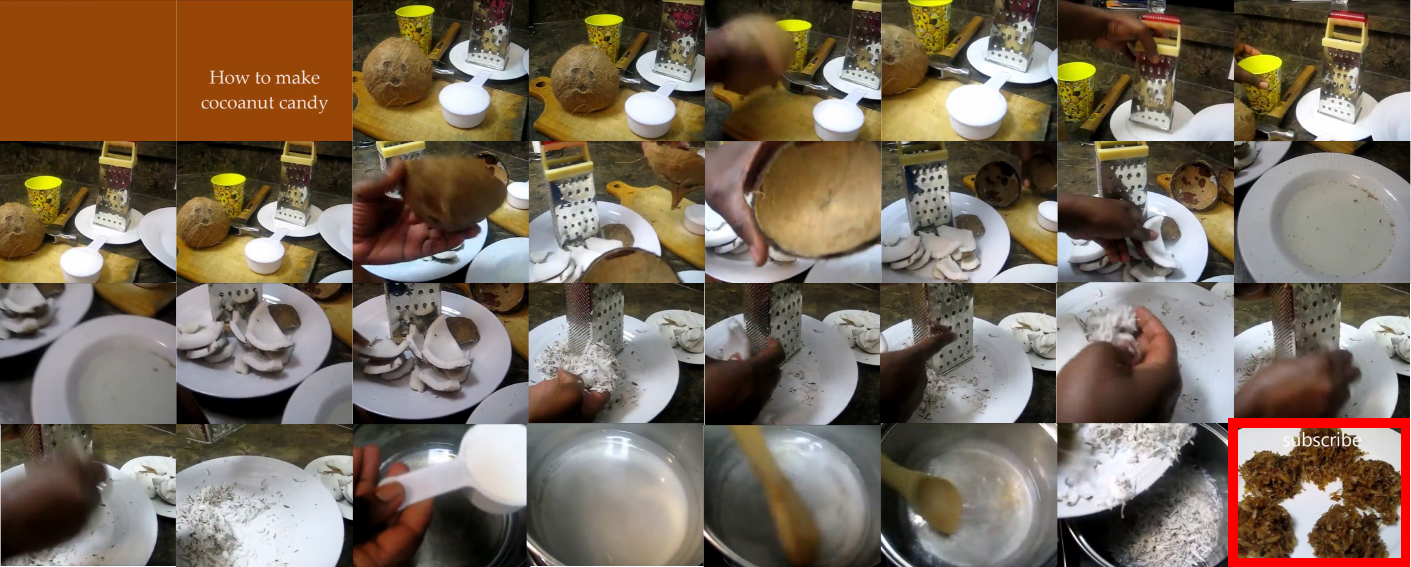}\vspace{5pt}} \\
\noalign{\vskip 0.25ex}\cdashline{1-2}\noalign{\vskip 0.75ex}
\textbf{Generated Answer} & \multicolumn{1}{p{.9\textwidth}}{To make coconut candy, start by grating the coconut using a box grater over a plate. Then, melt sugar in a pot and stir it until it turns into a caramel-like consistency. Add the grated coconut to the melted sugar mixture and stir well. Once the mixture is well combined, let it cool and solidify. Finally, shape the solidified mixture into small balls or shapes and enjoy your homemade coconut candy!} \\
\noalign{\vskip 0.25ex}\cdashline{1-2}\noalign{\vskip 0.75ex}
\textbf{ROUGE-L Score} & \multicolumn{1}{p{.9\textwidth}}{33.55} \\
\midrule
\midrule
\textbf{Ground Truth} & \multicolumn{1}{p{.9\textwidth}}{To make coconut candy squares, you'll need shredded coconut, white sugar, corn syrup, water, butter, salt, baking soda, and a small cake pan. Combine the sugar, water, and corn syrup in a saucepan over medium heat, stirring until the mixture boils. Then, add the butter, remove the mixture from the heat source, and stir in the remaining ingredients. Pour the mixture into the pan and let it set for a few hours. Then, just cut the solid into small squares and enjoy!} \\
\bottomrule

\end{tabular}
}
\label{tab:case_gen_coconut}
\vspace{-0.1in}
\end{table*}

\begin{table*}[t!]
\caption{Case study comparing \textsc{TextRAG (BM25)} and \textsc{VideoRAG-V} approaches.}
\vspace{-0.1in}
\scriptsize
\centering
\resizebox{\textwidth}{!}{
\renewcommand{\arraystretch}{0.5}
\begin{tabular}{l c}
\toprule
\multicolumn{2}{c}{\textbf{\textsc{\large TextRAG (BM25)}}} \\ 
\midrule
\textbf{Question} & \multicolumn{1}{p{.9\textwidth}}{Explain how to make a clay rose.} \\
\noalign{\vskip 0.25ex}\cdashline{1-2}\noalign{\vskip 0.75ex}
\textbf{Retrieved Document} & 
\multicolumn{1}{p{.9\textwidth}}{Vally Wiselthier, an art deco ceramicist who was making pieces for General Ceramics in Keasbey, New Jersey. After seeing her preliminary pieces, Erni gave Rose a membership at Greenwich House, where she learned how to use a potter's wheel. Shortly after their first son George was born, he was found to have intractable asthma. The Cabat's decided to move to Arizona around 1942 in order to alleviate his condition. During World War II Rose worked as a riveter at the Davis-Monthan Army Air Field repairing war-damaged aircraft. Rose was able to make primitive ceramics from the extra clay that Erni} \\ 
\noalign{\vskip 0.25ex}\cdashline{1-2}\noalign{\vskip 0.75ex}
\textbf{Generated Answer} & \multicolumn{1}{p{.9\textwidth}}{\textcolor{red}{The document does not provide a step-by-step guide on how to make a clay rose. However, based on the information provided, we can infer that Rose learned how to use a potter's wheel at Greenwich House after seeing Vally Wiselthier's preliminary pieces. This suggests that she may have received some training or instruction in pottery from Vally Wiselthier or other instructors at Greenwich House.}\newline\newline To make a clay rose, you would need to follow these general steps:\newline 1. Gather materials: You will need clay, a potter's wheel, a wire cutter, a sponge, and any additional materials you want to use for decoration (such as glaze, paint, or glitter).\newline 2. Prepare the clay: Cut a piece of clay to the desired size and shape for your rose. Use a wire cutter to trim any excess clay and smooth out any rough edges.\newline 3. Shape the clay: Use your hands or tools to shape the clay into a rose. You can start by creating a small ball of clay for the center of the rose, then add petals around it. Use a sponge to smooth out any rough areas and refine the shape of the rose.\newline 4. Add details: Use a tool or your fingers to add details to the petals, such as veins or shading. You can also use additional materials like glaze, paint, or glitter to decorate the rose.\newline 5. Allow the clay to dry: Let the clay dry completely before firing it in a kiln. This will help prevent any cracking or breaking during the firing process.\newline 6. Fire the clay: Place the dried clay rose in a kiln and fire it according to the manufacturer's instructions. This will harden the clay and create a durable piece of pottery.\newline 7. Glaze or finish: Once the clay is fired, you can apply a glaze or finish to give the rose a glossy or matte appearance. Allow the glaze to dry and cure before handling the finished piece.\newline\newline It's important to note that this is a general guide and the specific steps may vary depending on the type of clay you are using and the techniques you are employing. It's also important to follow safety guidelines when working with clay and kilns.} \\
\midrule
\multicolumn{2}{c}{\textbf{\large \textsc{\ours-V}}} \\
\midrule
\textbf{Question} & \multicolumn{1}{p{.9\textwidth}}{Explain how to make a clay rose.} \\
\noalign{\vskip 0.25ex}\cdashline{1-2}\noalign{\vskip 0.75ex}
\textbf{Retrieved Video} & \parbox[t]{.9\textwidth}{\vspace{-3pt}\includegraphics[width=.8\textwidth]{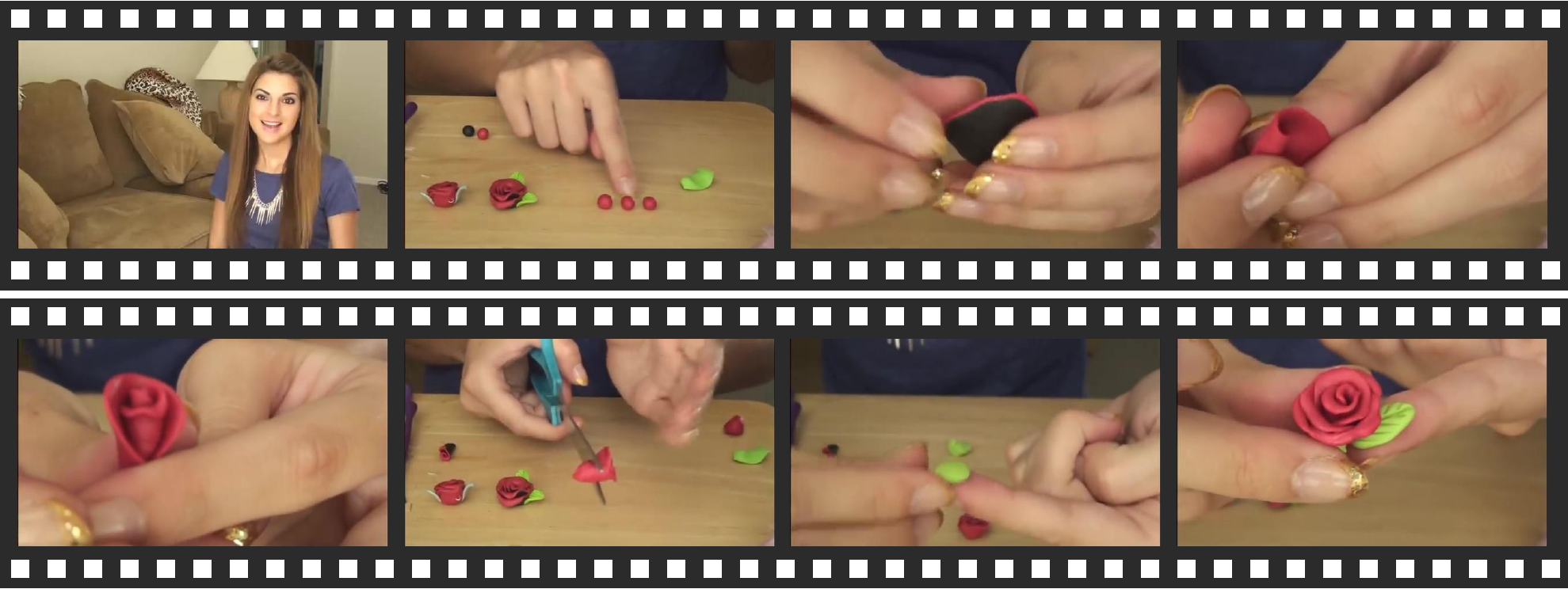}\vspace{5pt}} \\
\noalign{\vskip 0.25ex}\cdashline{1-2}\noalign{\vskip 0.75ex}
\textbf{Generated Answer} & \multicolumn{1}{p{.9\textwidth}}{To make a clay rose, start by rolling out a piece of clay into a flat sheet. Use a rolling pin to flatten the clay further. Then, use a knife or sculpting tool to cut the clay into petal shapes. Roll the petals into a spiral shape to form the rose. Use a small tool to create indentations in the petals for texture. Assemble the petals around a central stem, shaping them to resemble a rose. Add leaves and other details as desired. Finally, use a mold to imprint patterns onto the petals, adding depth and realism to the rose.} \\
\midrule
\midrule
\textbf{Ground Truth} & \multicolumn{1}{p{.9\textwidth}}{To make a clay rose, start by rolling a chunk of clay into a ball, dividing the ball into seven similarly-sized pieces, and rolling those pieces into balls. Take the smallest ball and press it into a guitar-pick-shaped \lq\lq petal'' with the pointy part thicker than the rounded part, then fold the sides gently inward. Repeat with a slightly bigger ball of clay, then fold this second petal around the first. Continue attaching slightly larger petals in a spiral, then pinch the bottom of your rose to connect all the petals.} \\
\bottomrule

\end{tabular}
}
\label{tab:qualitative1}
\vspace{-0.1in}
\end{table*}

\begin{table*}
    \caption{The prompt used for generating synthetic question-answer pairs.}
    \label{tab:prompt:synthetic}
    \vspace{-0.1in}
    \resizebox{0.95\textwidth}{!}{
    \renewcommand{\arraystretch}{1.1}
    \renewcommand{\tabcolsep}{2.5mm}
        \begin{tabular}{p{0.95\textwidth}}
        \toprule
        Your task is to create 3 diverse, relevant, and realistic question-answer pairs specifically designed to evaluate a Retrieval-Augmented Generation (RAG) system using the provided video. The questions should be crafted in a way that answering them requires retrieving the specific video or its information from a large corpus, without being overly specific or relying on minor details. Focus on crafting questions that are general enough to apply broadly yet detailed enough to leverage key information from the video. Avoid direct references such as 'in this video' or overly specific mentions that limit the question's scope to the given video. Instead, structure questions to include contextual cues or keywords that would aid in retrieving the correct content while maintaining natural language flow. \\
        \midrule
        Consider including questions that cover:\\
        - Generalized step-by-step actions or procedures (e.g., preparation steps, typical tasks) \\
        - Logical connections between steps (e.g., `What should be done after breaking apart the ingredients?') \\
        - Common tools or objects involved and their general purpose \\
        - Contextual or background details that support retrieval (e.g., setting or process clues) \\
        - Typical outcomes or results of observed actions or procedures \\
        
        \midrule
        
        The JSON structure should look like this: \\
        \texttt{[} \\
        \texttt{  \{``question'': ``<Insert Question 1>'', ``answer'': ``<Insert Answer 1>''\},} \\
        \texttt{  \{``question'': ``<Insert Question 2>'', ``answer'': ``<Insert Answer 2>''\},} \\
        \texttt{  \{``question'': ``<Insert Question 3>'', ``answer'': ``<Insert Answer 3>''\} } \\
        \texttt{]} \\
         ... up to 3 question-answer pairs \\
        \bottomrule
        \end{tabular}
    }
\end{table*}

\begin{table*}
    \caption{The prompt template used for G-Eval, which is further used as a guideline for human evaluation.}
    \label{tab:prompt:geval}
    \vspace{-0.1in}
    \renewcommand{\arraystretch}{1.2}
    \begin{tabular}{p{0.97\textwidth}} 
    \toprule
    You are tasked with evaluating a Generated Response to the given Question based on its overall quality compared to a provided Ground Truth Answer. 
    \\
    \midrule
    \textbf{Evaluation Criteria:} \\
    1. Carefully read the Ground Truth and the Generated Response. \\
    2. Assess how well the Generated Response matches the Ground Truth. Please penalize the Generated Response that has the far different content and style and is largely longer than the Ground Truth. \\
    3. Provide an overall score (1-5) based on your evaluation. \\
    \midrule
    Question: \texttt{\{\{Question\}\}} \\
    Ground Truth Answer: \texttt{\{\{Ground\_Truth\_Answer\}\}} \\
    Generated Response: \texttt{\{\{Generated\_Response\}\}} \\
    \midrule
    {Please provide only a single numerical rating (1, 2, 3, 4, or 5), without any additional commentary, formatting, or chattiness.} \\    
    \bottomrule
    \end{tabular}
\end{table*}

\end{document}